\documentclass{article}
\newif\ifemail
\newif\ifchecklist
\newif\ifappendix
\newif\ifbanner
\usepackage{soul}
\usepackage[small]{caption}
\usepackage{graphicx}
\usepackage{amsmath}
\usepackage{amsthm}
\usepackage{booktabs}
\usepackage{algorithm}
\usepackage{amssymb}
\usepackage{multirow}
\usepackage{wrapfig}
\usepackage{adjustbox}
\usepackage{makecell} 
\usepackage{colortbl}
\usepackage{bbding} %首先在导言区调用bbding包
\usepackage{color,xcolor}
\usepackage{subcaption}
\newtheorem{theorem}{Theorem}[section]

\usepackage{setspace}
\usepackage{url}
\usepackage{algorithmicx}
\usepackage{tikz}
\usepackage{mathtools}
\usepackage[normalem]{ulem}  %
\usepackage{tabu}
\usepackage{pgfplots,pgfplotstable}
\usepgfplotslibrary{fillbetween}
\usepackage{calc}
\usepackage{pbox}
\usepackage[noend]{algpseudocode}
\usepackage{afterpage}
\pgfplotsset{compat=1.14}	 %
\pgfplotsset{compat/show suggested version=false}
\usepackage[preprint,nonatbib]{neurips_2023}\emailfalse\checklistfalse\appendixtrue\bannertrue
\usepackage[numbers]{natbib}

\usepackage[utf8]{inputenc} % allow utf-8 input
\usepackage[T1]{fontenc}    % use 8-bit T1 fonts
\usepackage{hyperref}       % hyperlinks
\usepackage{url}            % simple URL typesetting
\usepackage{amsfonts}       % blackboard math symbols
\usepackage{nicefrac}       % compact symbols for 1/2, etc.
\usepackage{microtype}      % microtypography

\definecolor{olive}{rgb}{0.5, 0.5, 0.0}
\definecolor{maroon}{rgb}{0.69, 0.19, 0.38}
\definecolor{celestialblue}{rgb}{0.29, 0.59, 0.82}
\definecolor{darkgreen}{rgb}{0.0, 0.6, 0.0}
\definecolor{grey}{rgb}{0.5,0.5,0.5}
\definecolor{darkblue}{rgb}{0.19, 0.19, 0.62}
\definecolor{silver}{rgb}{0.7,0.7,0.7}
\definecolor{darkcyan}{rgb}{0.0, 0.55, 0.55}

\def\clap#1{\hbox to 0pt{\hss #1\hss}}%

\newcommand\undefcolumntype[1]{\expandafter\let\csname NC@find@#1\endcsname\relax}

\definecolor{C0}{rgb}{0.121569, 0.466667, 0.705882}
\definecolor{C1}{rgb}{1.000000, 0.498039, 0.054902}
\definecolor{C2}{rgb}{0.172549, 0.627451, 0.172549}
\definecolor{C3}{rgb}{0.839216, 0.152941, 0.156863}
\definecolor{C4}{rgb}{0.580392, 0.403922, 0.741176}
\definecolor{C5}{rgb}{0.549020, 0.337255, 0.294118}
\definecolor{C6}{rgb}{0.890196, 0.466667, 0.760784}
\definecolor{C7}{rgb}{0.498039, 0.498039, 0.498039}
\definecolor{C8}{rgb}{0.737255, 0.741176, 0.133333}
\definecolor{C9}{rgb}{0.090196, 0.745098, 0.811765}

\newcommand{\rr}[1]{\textcolor{red}{#1}}

\newcommand{\atphantom}{\vphantom{${}^2$}}
\newcommand{\AProcedure}[2]{\Procedure{\smash{#1}}{\smash{#2}}}
\newcommand{\AComment}[1]{\Comment{\smash{#1}}}
\newcommand{\AState}[1]{\State{\smash{#1}}}
\newcommand{\AFor}[1]{\For{\smash{#1}}}
\newcommand{\AIf}[1]{\If{\smash{#1}}}

\newcommand{\CUDone}{%
\begin{algorithm}[t]
\footnotesize
\captionof{algorithm}[CUD]{\atphantom\ \ The Training Procedure of Catch-Up Distillation (Runge-Kutta 12)}
\begin{spacing}{1.1}
\begin{algorithmic}[1]
  \AProcedure{CUD}{$f_\theta$,$g_{\psi 1}$, $X_0 \sim \pi_0$, $N\left(=50w\right)$, $\hat{h}=(1/16)$, $\alpha=(0.9999)$, $\epsilon=(1e\!-\!5)$, $\eta=(2e\!-\!4)$}  
      \AState{{\bf Initialize} $\theta_-\leftarrow\theta$, $\psi 1_-\leftarrow\psi 1$}
        \AComment{Initialize the EMA model}
      \AFor{$i \in \{1, \dots, N\}$}\AComment{Perform N iterations of training}
 \AState{{\bf sample} $t\sim\mathcal{U}[\epsilon,1]$, $h\sim\mathcal{U}[\epsilon,\hat{h}]$} 
       \AComment{Sampling time point and catch-up step size}
    \AState{{\bf reparameter} $X_1 \leftarrow q_\psi(X_0)$, {\bf compute} $X_t \leftarrow tX_1 + (1-t)X_0$}
        \AComment{Generate sample point $X_t$}
      \AState{$v_t \leftarrow g_{\psi 1}(f_\theta(X_t,t))$}
          \AComment{Calculate the velocity for the current time step}
    \AIf{$t-h\geq \epsilon$}  \AComment{Calculate the velocity for the next 1st time step}
      \AState{$v_{t-h} \gets g_{\psi 1}(f_\theta(\widetilde{X}_{t-h},t-h))$, $\widetilde{X}_{t-h} \gets X_t -v_th$} \AComment{The right distillation behaviour}
    \Else
          \AState{$v_{t-h} \gets X_1 - X_0$} \AComment{Avoid the wrong distillation behaviour}
    \EndIf
    \AState{$\mathcal{L}_\textrm{prior} = \beta\textrm{D}_{KL}(\textrm{Law}(X_1)||\pi_1)$} \AComment{Let the marginal distribution of $X_1$ approximate $\mathcal{N}(0,\textbf{I})$}
    \AState{$\mathcal{L}_\textrm{base} = \omega_1(i)\textrm{MSE}(v_{t-h},v_t)+\omega_2(i)\textrm{MSE}(X_1-X_0,v_t)$} \AComment{Our method's core loss function}
    \AState{$\theta \gets \theta - \eta \nabla_{\theta}\mathcal{L}_\textrm{CUD},\ \psi 1\gets\psi 1-\eta\nabla_{\psi 1}\mathcal{L}_\textrm{CUD},\ \psi\gets\psi-\eta\nabla_\psi\mathcal{L}_\textrm{CUD}$, $\mathcal{L}_\textrm{CUD} = \mathcal{L}_\textrm{prior}+\mathcal{L}_\textrm{base}$}
  \AState{{\bf update} $\theta_-\gets \alpha\theta_- + (1-\alpha)\theta,\psi 1_-\gets \alpha\psi 1_- + (1-\alpha)\psi 1$}\AComment{Update based on EMA}
    \EndFor
    \AState{\textbf{return} $\theta_-,\psi 1_-,\psi$}
        \AComment{Return the model parameters}
  \EndProcedure
\end{algorithmic}
\end{spacing}
\label{alg:CUDone}
\end{algorithm}
\vspace{-5pt}
}

\newcommand{\CUDtwo}{%
\begin{algorithm}[t]
\footnotesize
\captionof{algorithm}[CUD]{\atphantom\ \ The Training Procedure of Catch-Up Distillation (Runge-Kutta 23)}
\begin{spacing}{1.1}
\begin{algorithmic}[1]
  \AProcedure{CUD}{$f_\theta$,$\{g_{\psi i}\}_{i=1}^2$, $X_0 \sim \pi_0$, $N\left(=50w\right)$, $\hat{h}=(1/16)$, $\alpha=(0.9999)$, $\epsilon=(1e\!-\!5)$, $\eta=(2e\!-\!4)$}  
      \AState{{\bf Initialize} $\theta_-\leftarrow\theta$, $\{\psi j_-\leftarrow\psi j\}_{j=1}^2$}
        \AComment{Initialize the EMA model}
      \AFor{$i \in \{1, \dots, N\}$}\AComment{Perform N iterations of training}
 \AState{{\bf sample} $t\sim\mathcal{U}[\epsilon,1]$, $h\sim\mathcal{U}[\epsilon,\hat{h}]$} 
       \AComment{Sampling time point and catch-up step size}
    \AState{{\bf reparameter} $X_1 \leftarrow q_\psi(X_0)$, {\bf compute} $X_t \leftarrow tX_1 + (1-t)X_0$}
        \AComment{Generate sample point $X_t$}
      \AState{$v_t^1 \leftarrow g_{\psi 1}(f_\theta(X_t,t))$, $v_t^2 \leftarrow g_{\psi 2}(f_\theta(X_t,t))$}
          \AComment{Calculate the velocity for $g_{\psi 1}$, $g_{\psi 2}$}
      \AState{$v_{t-h}^\textrm{tmp} \leftarrow g_{\psi 1}(f_\theta(X_t-v_t^1h,t-h))$}
      \AState{}
          \AComment{Calculate the intermediate variables required by Runge-Kutta}
    \AIf{$t-2h\geq \epsilon$}  \AComment{Calculate the velocity for the next 1st and 2nd time step}
        \vspace{1mm}
      \AState{$\widetilde{X}_{t-2h} \gets X_t -(v_t^1+v_{t-h}^\textrm{tmp})h$, $\widetilde{X}_{t-h} \gets X_t -\frac{v_t^1+v_{t-h}^\textrm{tmp}}{2}h$} \AComment{The right distillation behaviour}
      \AState{$v_{t-2h} \gets g_{\psi 1}(f_\theta(\widetilde{X}_{t-2h},t-2h))$, $v_{t-h} \gets g_{\psi 1}(f_\theta(\widetilde{X}_{t-h},t-h))$} 
    \Else
          \AState{$v_{t-2h}\gets X_1-X_0$, $v_{t-h} \gets X_1-X_0$}\AComment{Avoid the wrong distillation behaviour}

    \EndIf
    \AState{$\mathcal{L}_\textrm{prior} = \beta\textrm{D}_{KL}(\textrm{Law}(X_1)||\pi_1)$} \AComment{Let the marginal distribution of $X_1$ approximate $\mathcal{N}(0,\textbf{I})$}
    \AState{$\mathcal{L}_\textrm{base} = \omega_1(i)\sum_{j=1}^2\textrm{MSE}(v_{t-jh},v_t^j)+\omega_2(i)\sum_{j=1}^2\textrm{MSE}(X_1-X_0,v_t^j)$} 
    \AState{} \AComment{Our method's core loss function}
    \AState{$\mathcal{L}_\textrm{CUD} =\mathcal{L}_\textrm{prior}+\mathcal{L}_\textrm{base}$}
    \AState{$\theta \gets \theta - \eta \nabla_{\theta}\mathcal{L}_\textrm{CUD},\ \{\psi j\gets\psi j-\eta\nabla_{\psi j}\mathcal{L}_\textrm{CUD}\}_{j=1}^2,\ \psi\gets\psi-\eta\nabla_\psi\mathcal{L}_\textrm{CUD}$}
  \AState{{\bf update} $\theta_-\gets \alpha\theta_- + (1-\alpha)\theta,\{\psi j_-\gets \alpha\psi j_- + (1-\alpha)\psi j\}_{j=1}^2$}\AComment{Update based on EMA}
    \EndFor
    \AState{\textbf{return} $\theta_-,\{\psi j_-\}_{j=1}^2,\psi$}
        \AComment{Return the model parameters}
  \EndProcedure
\end{algorithmic}
\end{spacing}
\label{alg:CUDtwo}
\end{algorithm}
}

\newcommand{\CUDthree}{%
\begin{algorithm}[!t]
\footnotesize
\captionof{algorithm}[CUD]{\atphantom\ \ The Training Procedure of Catch-Up Distillation (Runge-Kutta 34)}
\begin{spacing}{1.1}
\begin{algorithmic}[1]
  \AProcedure{CUD}{$f_\theta$,$\{g_{\psi i}\}_{i=1}^3$, $X_0 \sim \pi_0$, $N\left(=50w\right)$, $\hat{h}=(1/16)$, $\alpha=(0.9999)$, $\epsilon=(1e\!-\!5)$, $\eta=(2e\!-\!4)$}  
      \AState{{\bf Initialize} $\theta_-\leftarrow\theta$, $\{\psi j_-\leftarrow\psi j\}_{j=1}^3$}
        \AComment{Initialize the EMA model}
      \AFor{$i \in \{1, \dots, N\}$}\AComment{Perform N iterations of training}
 \AState{{\bf sample} $t\sim\mathcal{U}[\epsilon,1]$, $h\sim\mathcal{U}[\epsilon,\hat{h}]$} 
       \AComment{Sampling time point and catch-up step size}
    \AState{{\bf reparameter} $X_1 \leftarrow q_\psi(X_0)$, {\bf compute} $X_t \leftarrow tX_1 + (1-t)X_0$}
        \AComment{Generate sample point $X_t$}
      \AState{$v_t^1 \leftarrow g_{\psi 1}(f_\theta(X_t,t))$, $v_t^2 \leftarrow g_{\psi 2}(f_\theta(X_t,t))$, $v_t^3 \leftarrow g_{\psi 3}(f_\theta(X_t,t))$}
      \AState{}
          \AComment{Calculate the velocity for $g_{\psi 1}$, $g_{\psi 2}$, $g_{\psi 3}$}
      \AState{$v_{t-2h}^\textrm{tmp} \leftarrow g_{\psi 1}(f_\theta(X_t-\frac{7v_t^1+v_{t-h}^\textrm{tmp}}{4}h,t-2h))$, $v_{t-h}^\textrm{tmp} \leftarrow g_{\psi 1}(f_\theta(X_t-v_t^1h,t-h))$}
      \AState{}
          \AComment{Calculate the intermediate variables required by Runge-Kutta}
    \AIf{$t-3h\geq \epsilon$}  \AComment{Calculate the velocity for the next 1st, 2nd and 3rd time step}
    \vspace{1mm}
          \AState{$\widetilde{X}_{t-h} \gets X_t -\frac{5v_t^1+8v_{t-h}^\textrm{tmp}-v_{t-2h}^\textrm{tmp}}{12}h$}  \AComment{The right distillation behaviour}
        \vspace{3mm}
      \AState{$\widetilde{X}_{t-2h} \gets X_t -\frac{5v_t^1+8v_{t-h}^\textrm{tmp}-v_{t-2h}^\textrm{tmp}}{6}h$}
        \vspace{3mm}
      \AState{$\widetilde{X}_{t-3h} \gets X_t -\frac{5v_t^1+8v_{t-h}^\textrm{tmp}-v_{t-2h}^\textrm{tmp}}{4}h$} 
      \AState{$v_{t-h} \gets g_{\psi 1}(f_\theta(\widetilde{X}_{t-h},t-h))$} 
      \AState{$v_{t-2h} \gets g_{\psi 1}(f_\theta(\widetilde{X}_{t-2h},t-2h))$}
      \AState{$v_{t-3h} \gets g_{\psi 1}(f_\theta(\widetilde{X}_{t-3h},t-3h))$}
    \Else
          \AState{$v_{t-3h}\gets X_1-X_0$, $v_{t-2h}\gets X_1-X_0$, $v_{t-h} \gets X_1-X_0$}
          \AState{}\AComment{Avoid the wrong distillation behaviour}

    \EndIf
    \AState{$\mathcal{L}_\textrm{prior} = \beta\textrm{D}_{KL}(\textrm{Law}(X_1)||\pi_1)$} \AComment{Let the marginal distribution of $X_1$ approximate $\mathcal{N}(0,\textbf{I})$}
    \AState{$\mathcal{L}_\textrm{base} = \omega_1(i)\sum_{j=1}^3\textrm{MSE}(v_{t-jh},v_t^j)+\omega_2(i)\sum_{j=1}^3\textrm{MSE}(X_1-X_0,v_t^j)$} 
    \AState{} \AComment{Our method's core loss function}
    \AState{$\mathcal{L}_\textrm{CUD} =\mathcal{L}_\textrm{prior}+\mathcal{L}_\textrm{base}$}
    \AState{$\theta \gets \theta - \eta \nabla_{\theta}\mathcal{L}_\textrm{CUD},\ \{\psi j\gets\psi j-\eta\nabla_{\psi j}\mathcal{L}_\textrm{CUD}\}_{j=1}^3,\ \psi\gets\psi-\eta\nabla_\psi\mathcal{L}_\textrm{CUD}$}
  \AState{{\bf update} $\theta_-\gets \alpha\theta_- + (1-\alpha)\theta,\{\psi j_-\gets \alpha\psi j_- + (1-\alpha)\psi j\}_{j=1}^3$}\AComment{Update based on EMA}
    \EndFor
    \AState{\textbf{return} $\theta_-,\{\psi j_-\}_{j=1}^3,\psi$}
        \AComment{Return the model parameters}
  \EndProcedure
\end{algorithmic}
\end{spacing}
\label{alg:CUDthree}
\end{algorithm}
}

\newcommand{\ablationifuseema}{
\begin{figure*}[t]
\includegraphics[width=1\textwidth,trim={0cm 3cm 0cm 5cm},clip]{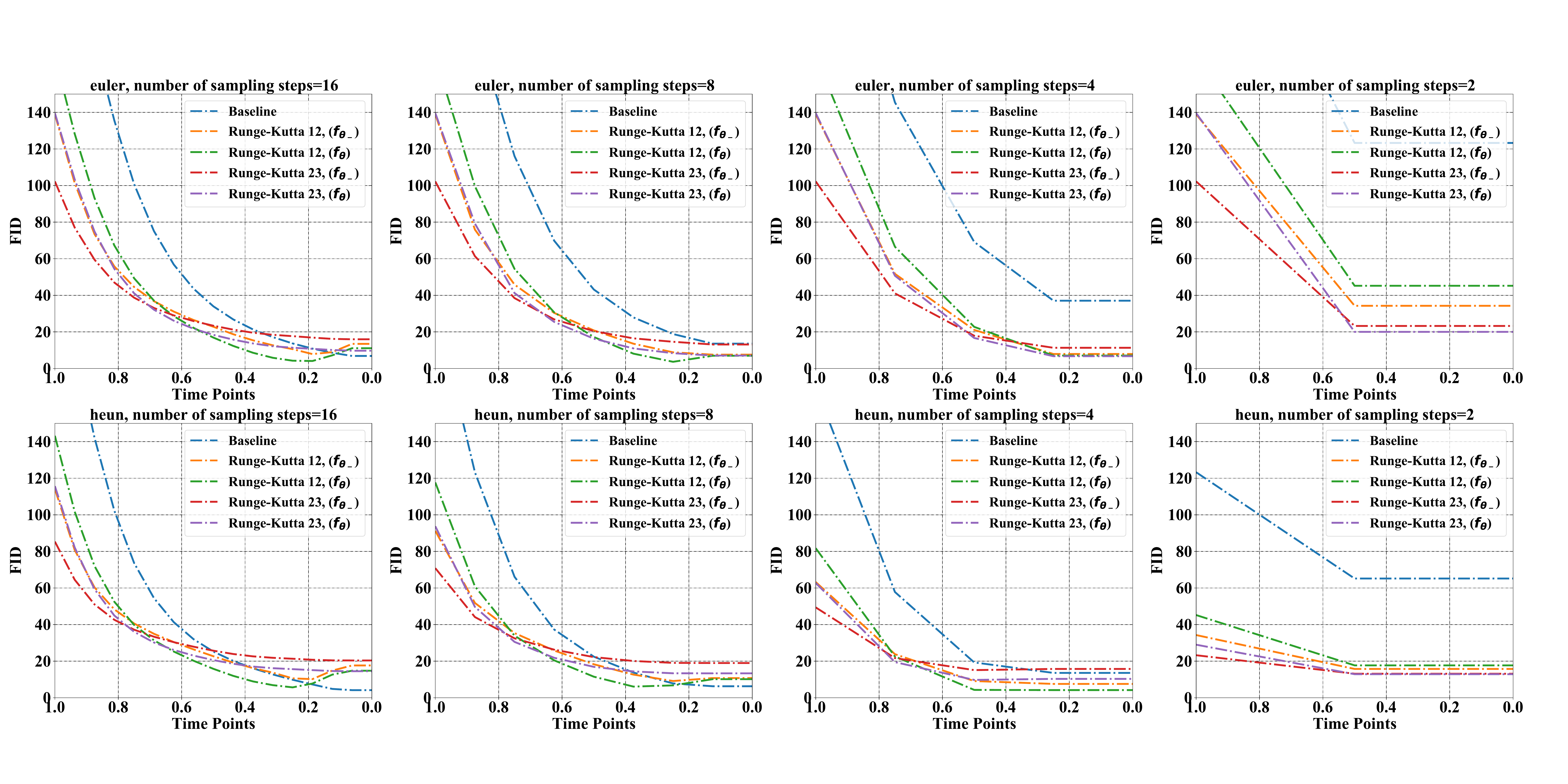}
\caption{Ablation experiments on whether to use the EMA model for CUD.}
\label{fig:ablation_study_if_use_ema}
\vspace{-9pt}
\end{figure*}
}

\newcommand{\ablationrungekutta}{
\begin{figure*}[t]
\includegraphics[width=1\textwidth,trim={0cm 0cm 0cm 2cm},clip]{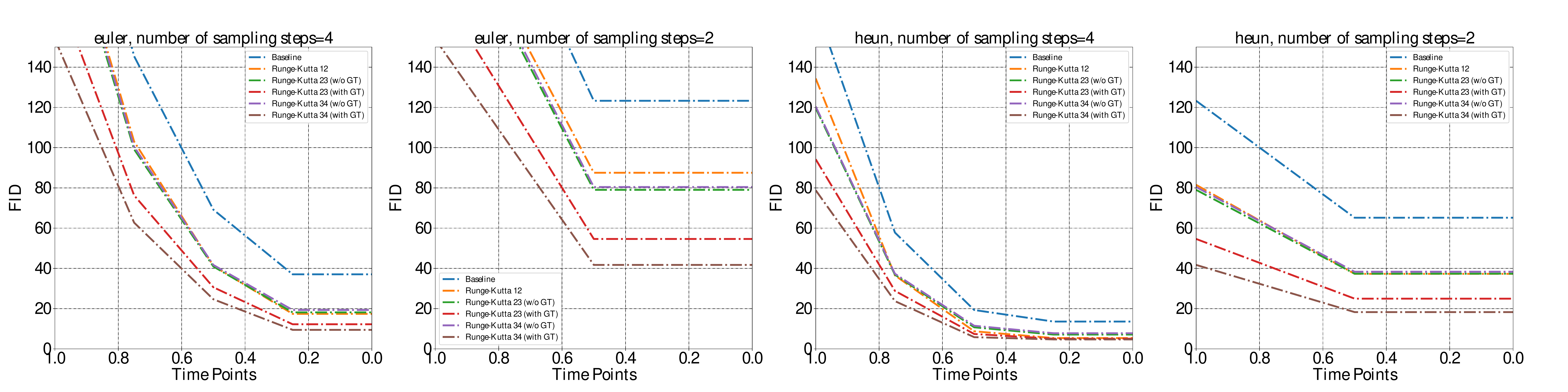}
\caption{Ablation experiments on Runge-Kutta-based multi-step alignment distillation.}
\label{fig:ablation_study_order_runge_kutta}
\vspace{-9pt}
\end{figure*}
}

\newcommand{\ablationrandomstepshakedrop}{
\begin{figure*}[!t]
\includegraphics[width=1\textwidth,trim={0cm 3cm 0cm 5cm},clip]{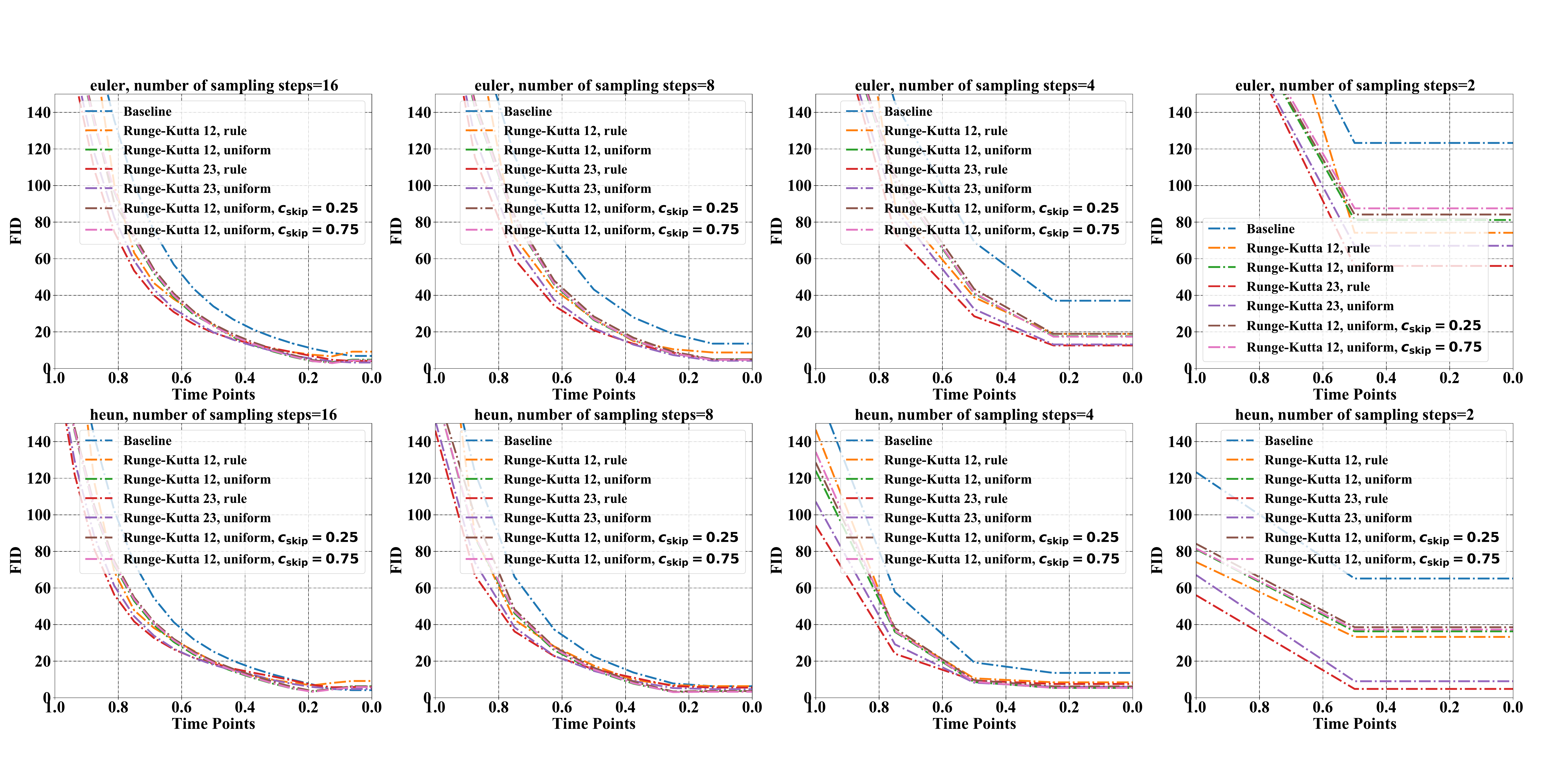}
\caption{Ablation experiments on whether to use the random step size and dynamic skip connection for CUD.}
\label{fig:ablation_study_if_use_random_step_and_shakedrop}
\vspace{-9pt}
\end{figure*}
}

\newcommand{\rowtl}[2]{\begin{tabular}[c]{@{}c@{}}#1\\#2\end{tabular}}
\title{Catch-Up Distillation: You Only Need to Train Once for Accelerated Sampling}

\author{Shitong Shao$^\dag$$^{\dag+}$, Xu Dai$^\dag$\thanks{Corresponding Author.}, Lujun Li$^{\dag-}$, Huanran Chen$^{\ddag}$, Yang Hu$^{\ddag *}$, Shouyi Yin$^\dag$ \\
$^\dag$Shanghai Artificial Intelligence Laboratory\\
$^{\dag+}$Southeast University, $^{\ddag}$Tsinghua University\\ 
$^{\dag-}$Hong Kong University of Science and Technology  \\
\texttt{1090784053sst@gmail.com};\quad \texttt{ daixu@pjlab.org.cn};\quad \texttt{yinsy@tsinghua.edu.cn} \\
\texttt{lilujunai@gmail.com};\quad \texttt{huanran\_chen@outlook.com};\quad\texttt{hu\_yang@tsinghua.edu.cn} \\
}

\begin{document}
\maketitle

\begin{abstract}
Diffusion Probability Models (DPMs) have made impressive advancements in various machine learning domains. However, achieving high-quality synthetic samples typically involves performing a large number of sampling steps, which impedes the possibility of real-time sample synthesis. Traditional accelerated sampling algorithms via knowledge distillation rely on pre-trained model weights and discrete time step scenarios, necessitating additional training sessions to achieve their goals. To address these issues, we propose the Catch-Up Distillation (CUD), which encourages the current moment output of the velocity estimation model ``catch up'' with its previous moment output. Specifically, CUD adjusts the original Ordinary Differential Equation (ODE) training objective to align the current moment output with both the ground truth label and the previous moment output, utilizing Runge-Kutta-based multi-step alignment distillation for precise ODE estimation while preventing asynchronous updates. Furthermore, we investigate the design space for CUDs under continuous time-step scenarios and analyze how to determine the suitable strategies. To demonstrate CUD's effectiveness, we conduct thorough ablation and comparison experiments on CIFAR-10, MNIST, and ImageNet 64$\times$64. On CIFAR-10, we obtain a FID of 2.80 by sampling in 15 steps under one-session training and the new state-of-the-art FID of 3.37 by sampling in one step with additional training. This latter result necessitated only 620k iterations with a batch size of 128, in contrast to Consistency Distillation, which demanded 2100k iterations with a larger batch size of 256. Our code is released at \rr{\textit{\url{https://anonymous.4open.science/r/Catch-Up-Distillation-E31F}.}}
\end{abstract}

\section{Introduction}
\vspace{-5pt}
Diffusion Probability Models (DPMs)~\cite{ddpm_begin,ddim,sde,nips2021_v_diffusion_model}, Variational Auto Encoders (VAEs)~\cite{vae_2,vae_3}, and Generate Adversarial Networks (GANs)~\cite{GAN_1,GAN_2} have achieved remarkable success across various applications, including image synthesis~\cite{nips2021_classifier_free_guidance}, audio synthesis~\cite{audio_1}, 3D reconstruction~\cite{iclr2023_dreamfusion}, and super-resolution~\cite{nc_sr_diffusion_model}. In recent years, DPMs, especially \textit{score-based} probabilistic models~\cite{sde}, have emerged as the new state-of-the-art family of generative models. They demonstrate superior abilities to generate more coherent and diverse samples compared to their VAE and GAN counterparts~\cite{iclr2021_guided_diffusion,yang2022diffusion}. This attribute to DPMs' theoretical completeness and exceptional image synthesis capabilities~\cite{sde,nips22_design,iclr22_rect}. \cite{sde} have shown that DPMs at continuous time steps can be interpreted as a \textit{score function} matching problem based on Stochastic Differential Equations (SDE) and Ordinary Differential Equations (ODE). Since the inception of this theory, research endeavors including SNIPS~\cite{nips21_b5c01503} and Analytic-DPM~\cite{iclr22_analytic} have probed into SDE-based generative models, amplifying their applicability. The stochastic sampling employed by SDE-based models elevates the quality of synthetic samples over deterministic sampling. Nevertheless, these advanced designs also involve expensive computational budgets, posing significant challenges to resource-constrained research labs and practical applications in industry.

As a result, researchers have also explored new paradigms that incorporate SDE-based training and ODE-based sampling, which have effectively accelerated sampling as demonstrated in studies such as ~\cite{ddim,dpm_solver,dpm_solver++}. Unfortunately, this paradigm is limited by the bottleneck of being \textit{training-free} (\textit{i.e.}, can be directly served for acceleration in inference without training overhead), resulting in weaker accelerated sampling effects compared to \textit{training-dependent} (\textit{i.e.}, need additional training overhead to achieve accelerated sampling) paradigms. A popular and widely-used \textit{training-dependent} accelerated sampling paradigm involves the application of knowledge distillation to expedite the sampling process~\cite{vanillakd,kdsurvey}. This paradigm was first proposed in Progressive Distillation (PD)~\cite{iclr22_progressive}, with the core idea being to achieve accelerated sampling incrementally by distilling a multi-step process into a single step. After that, a range of works~\cite{arxiv21_kd_diffusion,cvpr22_kd_guided,sun2022accelerating,icml23_consistency} have been carried out to improve PD's performance or extend its application scenarios. Among them,~\cite{sun2022accelerating} additionally distills the intermediate layer output of the noise estimation model, and~\cite{cvpr22_kd_guided} extends the PD algorithm to conditional sampling scenarios. Just recently, Consistency Distillation~\cite{icml23_consistency} was proposed to achieve the goal of the PD algorithm through a single additional training session, as opposed to multiple progressive training sessions.

However, all currently proposed distillation-based accelerated sampling algorithms have the following drawbacks: \textbf{(a)} They involve one or more additional training stages for distilling. \textbf{(b)} They need pre-training weights, and it must be ensured that the student's architecture is identical to that of the teacher. \textbf{(c)} They only consider discrete time steps and thus cannot synthesize images taking advantage of arbitrary numerical integration algorithms (\textit{e.g.}, Euler–Maruyama method and Runge-Kutta method). These shortcomings have, to some extent, prevented the application of these distillation-based accelerated sampling algorithms to broader DPM paradigms. To address these issues, we propose Catch-Up Distillation (CUD), which treats the previous moment output of the velocity estimation model as the teacher output and the current moment output of the velocity estimation model as the student output, and applies Runge-Kutta-based multi-step alignment distillation (as illustrated in Fig.~\ref{fig:cud_overall_structure}) to let the student output ``catch up'' with the teacher output while aligning the student output with the ground truth label. Based on this, CUD can complete accelerated sampling with a single training session, \textit{i.e.}, the original training session of DPMs, without requiring any pre-trained weights. Our contribution can be summarized as follows:
\begin{figure*}[t]
\includegraphics[width=1.\textwidth,trim={0cm 0cm 0cm 0cm},clip]{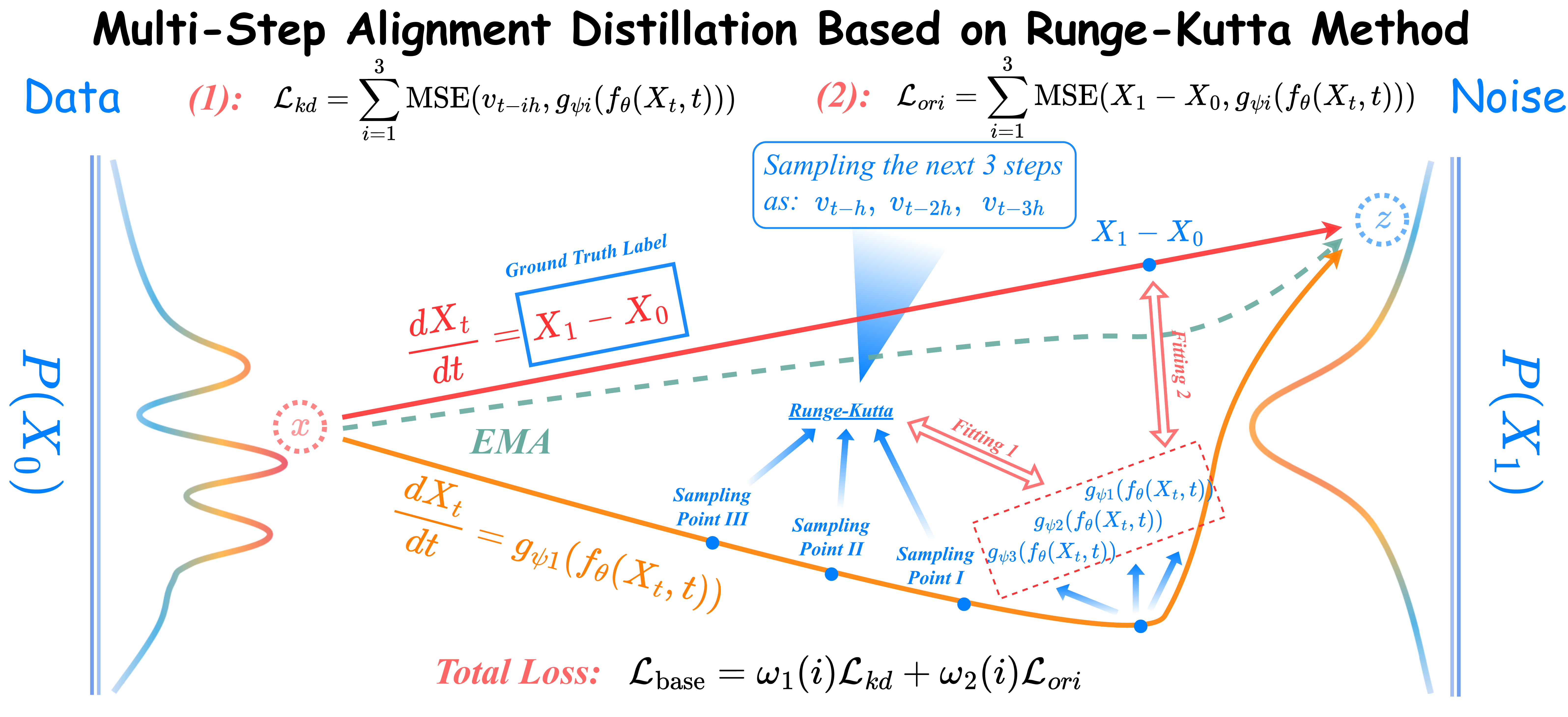}
\caption{The framework of Runge-Kutta-based multi-step alignment distillation (Runge-Kutta 34).}
\label{fig:cud_overall_structure}
\vspace{-11pt}
\end{figure*}
\begin{itemize}
\setlength{\itemsep}{0pt}
\setlength{\parsep}{0pt}
\setlength{\parskip}{0pt}
\item We present Catch-Up Distillation (CUD), the first accelerated sampling framework in a single training session without pre-training weights to extend the applicability of distillation-based accelerated sampling algorithms.
\item We search the design space of CUD and obtain some suitable strategies based on experiments and theories, which allow for a significant improvement in the quality of the synthetic samples.
\item We conduct extensive comparison and ablation experiments on the CIFAR-10, MNIST, and ImageNet 64$\times$64 datasets to verify that CUD can achieve superior performance compared to the original DPM training, using the same number of iteration steps.
\end{itemize}
\section{Background}
% Insert the procedure of catch-up distillation
\paragraph{Rectified flow.} It is well known that ODE-based training and sampling are conceptually simpler and require fewer Network Function Evaluations (NFEs) compared to those based on SDEs. Thus, in this work, we focus on \textit{ODE-based} generative models and design a novel algorithm with the expectation that our CUD can be done in a single training session without the need for pre-trained weights, ultimately achieving accelerated sampling. Rectified flow~\cite{iclr22_rect} is a recent work that achieves ODE-based training by minimizing the transport cost between marginal distributions $\pi_1$ and $\pi_0$, \textit{i.e.}, $\mathbb{E}[c(X_1-X_0)]$, where $X_1 = \textrm{Law}(\pi_1)$, $X_0 = \textrm{Law}(\pi_0)$ and $c:\mathbb{R}^d \rightarrow \mathbb{R}$ refers to a cost function. Due to the computational complexity of Optimal Transport (OT), Rectified flow provides a straightforward yet effective approach to generate a new coupling from a given one, which can be optimized using Stochastic Gradient Descent (SGD), a widely adopted optimization method in deep learning:
\begin{equation}
\small
\begin{aligned}
\theta^* = \operatorname*{arg\,min}_{\theta}\mathbb{E}_{t\sim \mathcal{U}[\epsilon,1]}\mathbb{E}_{X_0,X_1 \sim \pi_0,\pi_1} [\lambda(t)\textrm{MSE}(X_1-X_0 - f_\theta(X_t,t))],
\end{aligned}
\label{eq:standard_ode}
\end{equation}
where $\lambda(t)$ and $\textrm{MSE}$ refer to the weight function that satisfies $\lambda(t)\equiv1$ and Mean Square Error (MSE), respectively. $\epsilon$ is a very small amount (\textit{i.e.} 1e-5) to empirically avoid unnecessary fitting overhead. Furthermore, $X_t=tX_1+(1-t)X_0$ is chosen to ensure that $\forall f_\theta(X_t,t)$ fits the same target velocity $X_1-X_0$. After training the sampling can be done by \textit{definite integral}, \textit{i.e.}, $Z_0 = Z_1 + \int_{1}^{0} f_{\theta^*}(Z_\tau,\tau)d\tau$, \textit{s.t.}, $Z_1 = X_1$. Intuitively, we can utilize this ODE solution approach to gradually reduce noise in a ``clean'' image, and restore it to the original data distribution, denoted as $\pi_0$. In a series of well-established studies~\cite{dpm_solver,dpm_solver++,ddim,iclr22_rect,nips22_design}, numerical methods such as Euler's method, Heun's method, and Runge-Kutta method, have been effectively employed for solving ODEs in the sampling phase.

\paragraph{Reparameterized Noise Encoder.} Rectified flow employs a distillation technique to reduce transport costs by fitting $Z_1\!-\!Z_0$ instead of $X_1\!-\!X_0$. Commonly, this method requires an additional training phase. Do not like vanilla Rectified flow, in~\cite{icml23_curvature}, the authors propose defining a reparameterized noise encoder $q_\psi(\widetilde{X}_1|X_0)$ to reparameterize Gaussian noise, ensuring a smooth mapping from $X_0$ to $X_1$ and inducing efficient optimization. The new optimization objective can be expressed as
\begin{equation}
\small
\begin{aligned}
\operatorname*{arg\,min}_{\theta,\psi}\mathbb{E}_{t\sim \mathcal{U}[\epsilon,1]}\mathbb{E}_{\widetilde{X}_1 = q_\psi(X_0), X_0\sim \pi_0} [\lambda(t)\textrm{MSE}(\widetilde{X}_1-X_0 - f_\theta(X_t,t))+ \beta \textrm{D}_{KL}(q_\psi(\widetilde{X}_1|X_0)||\pi_1)].
\end{aligned}
\label{eq:curvature_ode}
\end{equation}
Here, $\beta$ and $\textrm{D}_{KL}$ represent the loss weight (default as $20$) and Kullback-Leibler divergence, respectively. Although $q_\psi(z|x)$ is primarily designed to minimize the \textit{curvature} on the transport path, we demonstrate that it also reduces transport costs empirically. We substantiate this through Theorem~\ref{the:law_1}, which implies that as the Mutual Information (MI) between two distributions increases, the corresponding transport cost diminishes. Higher MI signifies a stronger correlation between the two probability distributions. This means that as long as the cost function chosen for training is inversely proportional to MI, we can observe a smaller transport cost between samples drawn from these distributions than without a reparameterized noise encoder.
\begin{theorem}
\label{the:law_1}
(Proof in Appendix~\ref{apd:transport_cost}) If the cost function $\mathbb{E}[c(a-b)]\propto 1/I(\textrm{Law}(a),\textrm{Law}(b))$, where $a$ and $b$ are samples drawn from any pair of marginal distributions, then the transportation cost between $\textrm{Law}(\widetilde{X}_1)$ and $\textrm{Law}({X}_0)$ is smaller than the transportation cost between $\textrm{Law}({X}_1)$ and $\textrm{Law}({X}_0)$.
\end{theorem}
With the above analysis in mind, we will take the optimization objective~\ref{eq:curvature_ode} as a baseline and explore how to utilize distillation for accelerated sampling.

\paragraph{Knowledge Distillation in Accelerated Sampling.} \textit{Knowledge distillation} can accelerate the sampling process of DPMs. The studies of~\cite{iclr22_progressive,cvpr22_kd_guided,sun2022accelerating,arxiv21_kd_diffusion,icml23_consistency} compress the multi-step sampling process of the DPM into a single step, effectively reducing computational overhead without compromising the quality of the synthetic images. In particular,~\cite{icml23_consistency} proposes Consistency Distillation (CD), which is capable of superior image generation in one-step sampling scenarios. Specifically, CD first initializes the weights of the student model from the pre-trained weights, and makes the weights of the teacher model updated from Exponential Moving Average (EMA), and then lets the output of the teacher model at previous moment to supervise the output of the student model at the current moment, and finally obtains a distilled student model. However, these algorithms require pre-trained weights, and the student architecture is heavily dependent on the selected teacher architecture, limiting their scalability. In this paper, we aim to enable a DPM to function as both a teacher and a student, and accelerate sampling within a single training session without requiring any pre-trained weights, as compared to the above-mentioned work. It means that CUD can be considered a standard training session for DPMs with some additional loss terms, therefore ensuring its good portability.
\section{Methodology}
The conventional training objective for \textit{ODE/SDE-based} generative models is to take input samples $X_t$ at different time points $t$ and output the noise $X_1$, sample $X_0$, or velocity $X_1\!-\!X_0$, aligning them with the corresponding ground truth labels. This paradigm is incapable of allowing the model to generate high-quality samples in a few sampling steps scenario. Although Song \textit{et.al.} propose Consistency Training (CT), the method is only applicable to \textit{empirical PF ODE}~\cite{icml23_consistency} since it needs Karra's diffusion model paradigm~\cite{nips22_design} for supervision (detailed explanation can be found in Appendix~\ref{apd:cdt}) and has a huge performance gap compared to their proposed CD. These shortcomings prevent CT from generalizing to other diffusion model paradigms, \textit{e.g.}, VP-SDE, VE-SDE. Considering this issue, our proposed Catch-Up Distillation (CUD) aims to adjust the original training paradigm so that it not only performs ground truth label alignment but also enables $f_\theta(X_t,t)$ ``catch up'' with the output from \textit{catch-up sampling}. The term ``\textit{catch-up sampling}'' refers to using a numerical integral solver to estimate $\widetilde{X}_{t-h}$ from $X_t$, where ``$h$'' refers to the step size of the discrete sampling. As presented in Fig.~\ref{fig:cud_overall_structure}, CUD leverages Runge-Kutta-based multi-step alignment distillation for achieving accelerated sampling. Particularly, CUD also includes a series of simple but effective strategies (\textit{e.g.} use the training model for \textit{catch-up sampling}, random step size, dynamic skip connection) derived from searching the design space. Ultimately, the procedures of the CUD algorithm using Runge-Kutta 12, 23, and 34 are presented in Algorithms~\ref{alg:CUDone},~\ref{alg:CUDtwo} and~\ref{alg:CUDthree}, respectively. Algorithms~\ref{alg:CUDtwo} and~\ref{alg:CUDthree} and CUD's limitations and broader impact can be found in the Appendix~\ref{sec:rkb_ms_ada} and~\ref{sec:limitation}.

\CUDone
\subsection{Basic Catch-Up Distillation}
Integrating CD into continuous time steps may seem beneficial, but as demonstrated in Table~\ref{tab:ablation_study_result_1} in our experiments, it can lead to training collapse. This issue arises because the EMA model, $f_{\theta_-}(\cdot,\cdot)$, which guides training, may become unreliable due to inaccurate updates. {We can derive Theorem~\ref{the:swa_1} and thus simply explicate this fact.
\begin{theorem}
\label{the:swa_1}
(Proof in Appendix~\ref{apd:the_working_principle}) Assume that $\operatorname*{sup}_{X_t} L(f_\theta(X_t,t),f_{\theta_-}(X_t-hf_{\Psi}(X_t,t),t-h))\leq \gamma_1$ and $\operatorname*{sup}_{X_t} L(f_\theta(X_t,t),f_{\theta_-}(X_t,t))\leq \gamma_2$ after training convergence, where $L$, $\theta_-$, and $\Psi$ denote the norm, the parameters updated via EMA, and the pre-trained weight, respectively. And $f_\theta(\cdot,t)$ satisfies Lipschitz condition, \textit{i.e.}, $||f_\theta(x,t)-f_\theta(y,t)|| \leq K ||x-y||$, there exists $K>0$ such that for all $t \in [\epsilon,1]$. Then if $\epsilon \leq t_a \leq t_b \leq 1$ exists, we can obtain that
\begin{equation}
\small
\begin{aligned}
&||f_\theta(X_{t_b},t_b) - f_\theta(X_{t_a},t_a)|| \leq  \frac{t_b-t_a}{h}\left[\gamma_1+\gamma_2+Kh||f_\Psi(X_{t_b},t_b)-(X_1-X_0)||\right]. \\
\end{aligned}
\label{eq:why_core_function_work}
\end{equation}
\end{theorem}
The theorem establishes an upper bound, ensuring the stability of the distillation process. When the EMA model's one-step update is inaccurate, $\gamma_2$ will become larger, leading CD to training collapse. It also implies that the difference between $f_\Psi(X_{t_b},t_b)$ and $X_1-X_0$ must be adequately small for the process to remain stable. In the standard diffusion model training session, we do not have a pre-trained weight like CD, so we have to replace $\Psi$ with $\theta$. In line with common distillation algorithms, such as vanilla KD~\cite{vanillakd}, RKD~\cite{RKD}, and HSAKD~\cite{HSAKD}, the teacher model output and the ground truth label mutually supervise the student model output. This approach enhances the student model's generalization ability empirically. Consequently, we introduce a ground truth label to enable effective supervision, thereby avoiding the EMA model's misdirected guidance. The output of $f_\theta(X_t,t)$, for all $t \in [\epsilon,1]$, should align with $X_1 - X_0$. We term this minimal functional algorithm as basic CUD, which synchronizes the ground truth label while ``catches up'' to the model output at the previous moment. Thus, the new base loss can be denoted as\footnote{We omit the optimization of $\psi$ in all subsequent equations but utilize it in our experiments.}
\begin{equation}
\small
\begin{aligned}
& \mathcal{L}_{\textrm{base}}=\mathbb{E}_{t\sim \mathcal{U}[\epsilon,1]}\mathbb{E}_{\widetilde{X}_1 = q_\psi(X_0), X_0\sim \pi_0} [\omega_1(i) L(\widetilde{X}_{t-h},f_\theta(X_t,t)) +\omega_2(i)L(\widetilde{X}_1-X_0,f_\theta(X_t,t))], \\
\end{aligned}
\label{eq:base_loss}
\end{equation}
where $X_t = t\widetilde{X}_1+(1-t)X_0$ and $\widetilde{X}_{t-h}$ is calculated by \textit{catch-up sampling} using ODE $\frac{dX_t}{dt}=f_{\theta_-}(X_t,t)$. And $\omega_1,\omega_2$ represent the loss weights, which can even be parameterized by the number of current iterations, \textit{i.e.}, \textbf{\textit{dynamic weight}} in Table~\ref{tab:ablation_study_result_1}. 

\subsection{Runge-Kutta-Based Multi-Step Alignment Distillation}
One of the key components of our CUD is \textit{catch-up sampling}, which can be solved by different ODE solvers. Notably, Euler's method and Heun's method are the subsets of Runge-Kutta methods, \textit{w.r.t.}, Runge-Kutta 12, and Runge-Kutta 23. So in this study, we apply Runge-Kutta methods to model \textit{catch-up sampling} on a generic perspective. We only consider the Runge-Kutta algorithm up to order 3, as higher-order algorithms would lead to excessive computational overhead, even though it is possible to ignore the backpropagation of gradients and parameter updates by means of inference form, \textit{e.g.}, \texttt{torch.no\_grad()}. In general, the points sampled\begin{wrapfigure}{r}{5.2cm}
\includegraphics[height=0.2\textwidth,trim={0.1cm 0cm 0cm 0cm},clip]{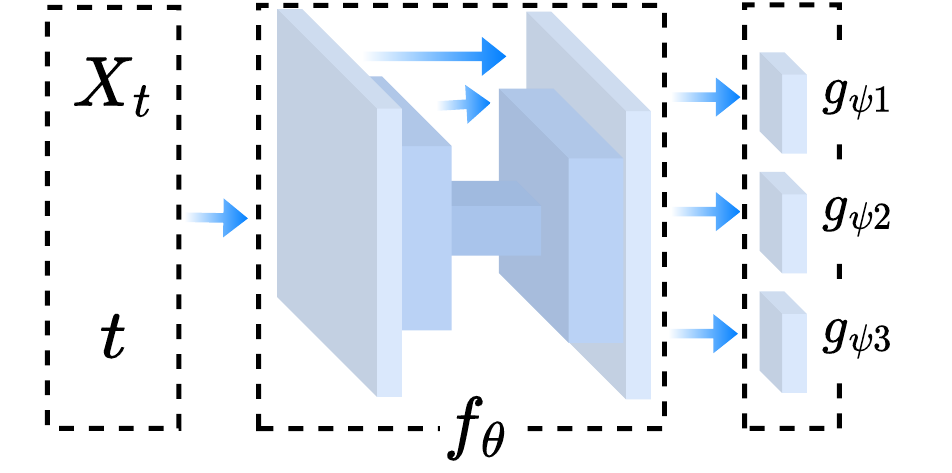}
\caption{The architecture of the velocity estimation model with various heads.}
\label{fig:architecture_of_unet}
\vspace{-10pt}
\end{wrapfigure} by the higher-order Runge-Kutta will only serve for one sampling step. But this format obviously wastes a very large amount of training overhead, so is it possible to make the best use of all the sampling points? The answer is yes, as shown in Fig.~\ref{fig:cud_overall_structure}. We can achieve Runge-Kutta-based multi-step alignment distillation, which aims to give the next step, the next-next step and even the next-next-next step of the estimated velocities simultaneously through all available sampling points, and then align them with outputs of different heads $\{g_{\psi i}(\cdot)\}_{i=1}^n$, where $n$ refers to the order of the Runge-Kutta method. \textbf{Note that $g_{\psi 1}(f_\theta(\cdot,\cdot))$ here is equivalent to $f_\theta(\cdot,\cdot)$ as mentioned earlier}. As illustrated in Fig.~\ref{fig:architecture_of_unet}, the input of all heads is the output of $f_\theta(\cdot,\cdot)$, and all heads are single meta-encoders that can be modeled as the sequence of consecutive \texttt{GroupNorm-SiLU-Conv}. By derivation in Appendix~\ref{apd:derivation_rk_msad}, we can derive Runge-Kutta (23/34)-based multi-step alignment distillation as follows:
\begin{equation}
\small
\begin{aligned}
& k_1=g_{\psi 1}(f_\theta(X_t,t)),k_2=g_{\psi 1}(f_\theta(X_t-hk_1,t-h)), \\
& \widetilde{X}_{t-h} = X_t - [\frac{1}{2}hk_1+\frac{1}{2}hk_2],\widetilde{X}_{t-2h} = X_t - [hk_1+hk_2], \\
& v_{t-h} = g_{\psi 1}(f_\theta(\widetilde{X}_{t-h},t-h)), v_{t-2h} = g_{\psi 1}(f_\theta(\widetilde{X}_{t-2h},t-2h)), \\
&\mathcal{L}_{\textrm{base}} = \omega_1(i)\sum_{i=1}^2\textrm{MSE}(v_{t-ih},g_{\psi i}(f_\theta(X_t,t))) + \omega_2(i)\sum_{i=1}^2\textrm{MSE}(\widetilde{X}_1-X_0,g_{\psi i}(f_\theta(X_t,t))),\quad \textcolor{C3}{\#\textrm{Runge-Kutta 23}} \\
\end{aligned}
\nonumber
\end{equation}
\vspace{-4pt}
\begin{equation}
\small
\begin{aligned}
& k_1=g_{\psi 1}(f_\theta(X_t,t)),k_2=g_{\psi 1}(f_\theta(X_t-hk_1,t-h)),k_3=g_{\psi 1}(f_\theta(X_t-\frac{7}{4}hk_1-\frac{1}{4}hk_2,t-2h)), \\
& \widetilde{X}_{t-h}\! =\! X_t\!-\![\frac{5}{12}hk_1\!+\!\frac{2}{3}hk_2\!-\!\frac{1}{12}hk_3],\widetilde{X}_{t-2h}\!=\!X_t\!-\![\frac{5}{6}hk_1\!+\!\frac{4}{3}hk_2\!-\!\frac{1}{6}hk_3],\widetilde{X}_{t-3h}\! =\! X_t\!-\![\frac{5}{4}hk_1\!+\!2hk_2\!-\!\frac{1}{4}hk_3], \\
& v_{t-h} = g_{\psi 1}(f_\theta(\widetilde{X}_{t-h},t-h)), v_{t-2h} = g_{\psi 1}(f_\theta(\widetilde{X}_{t-2h},t-2h)), v_{t-3h} = g_{\psi 1}(f_\theta(\widetilde{X}_{t-3h},t-3h)) , \\
&\mathcal{L}_{\textrm{base}} = \omega_1(i)\sum_{i=1}^3\textrm{MSE}(v_{t-ih},g_{\psi i}(f_\theta(X_t,t)))+ \omega_2(i)\sum_{i=1}^3\textrm{MSE}(\widetilde{X}_1-X_0,g_{\psi i}(f_\theta(X_t,t))),\quad \textcolor{C3}{\#\textrm{Runge-Kutta 34}}  \\
\end{aligned}
\label{eq:runge_kutta_distillation}
\end{equation}
Compared to a simple one-step alignment distillation, the use of multi-step alignment distillation provides more comprehensive information about $\frac{X_t}{dt} = g_{\psi 1_-}(f_{\theta_-}(X_t,t))$, preventing asynchronous model updates and improving model performance.

\subsection{Investigating Design Space}
\label{sec:design_space}
To some extent, the basic CUD has facilitated accelerated sampling; however, further enhancements are feasible. In this subsection, we delineate the design space and conduct a comprehensive analysis, ultimately proposing strategies superior to the basic CUD.

% For example, the loss function~\ref{eq:base_loss} could be improved, as the estimation of the EMA model actually leads to inaccurate results, \textit{i.e.}, the term $\gamma_2$ in Theorem~\ref{the:swa_1}. Furthermore, the quality of samples generated using more steps decreased compared to those generated using fewer steps. Additionally, the quality of samples generated using higher-order ODE solvers decreased compared to those generated using lower-order solvers. We point out that this is a result of the fixed step size $h$ of \textit{catch-up sampling} in ODE. Finally, different $t$ requires different fitting costs (\textcolor{red}{see Appendix~\ref{apd:cost_fo_fit_analysis}}), so designing a more plausible velocity estimation model still needs to be explored.

\paragraph{Limitations of Using EMA Model in \textit{Catch-Up Sampling}.} The update of the EMA model can enhance the model's generalization ability. Nevertheless, during the training phase, the disparity between the parameters of the EMA model and the training model may cause the upper bound of Theorem~\ref{the:swa_1} to be excessively loose, particularly in relation to the term $\gamma_2$ (as concluded in Appendix~\ref{apd:the_working_principle}). A more sensible and effective approach is to discard the EMA model and utilize the training model directly for \textit{catch-up sampling}. In this case, the errors introduced by the incorrect estimation of the EMA model will no longer exist, further improving the performance of the model.

% Some of the results in Table~\ref{tab:ablation_study_result_1} and Figs.~\ref{fig:ablation_study_if_use_ema} are still unsatisfactory, \textit{e.g.}, the results obtained by sampling with Heun's method are worse than those obtained by sampling with Euler's method. 

\paragraph{Random Step Size in \textit{Catch-Up Sampling}.} For accelerated sampling, a large interval between sample points is employed to reduce the sample count. However, in widely-used diffusion models such as DDPM~\cite{ddpm_begin}, NCSN~\cite{song2019generative}, and Rectified flow, the interval between sampling points approaches zero, as these models necessitate accurate estimation of differential equations to prevent training collapse. Intuitively, utilizing a fixed value of $h$ for CUD may result in the model converging to suboptimal solutions. This occurs because, for $X_t$, the model can only obtain the solution in the time point $t-h$ and is unable to explore solutions in other time points that could provide a more precise estimate of the ODE. To tackle this problem, we implement a non-fixed-step size strategy, \textit{uniform}, implying that the new step size $h$ follows a uniform distribution $\mathcal{U}[\epsilon,\hat{h}]$ ($\hat{h}$ is a predetermined step size, defaulting to 1/16). Additionally, we introduce a strategy named \textit{rule}, which determines the new step size as $h = t\hat{h}$, aiming to ensure the quality of the synthetic image retains a degree of ambiguity when $t\rightarrow 1$, necessitating a larger step size to augment the distillation strength. Conversely, when\begin{wrapfigure}{r}{6.2cm}\vspace{-10pt}
\includegraphics[height=0.2\textwidth,trim={0.1cm 0cm 0cm 0cm},clip]{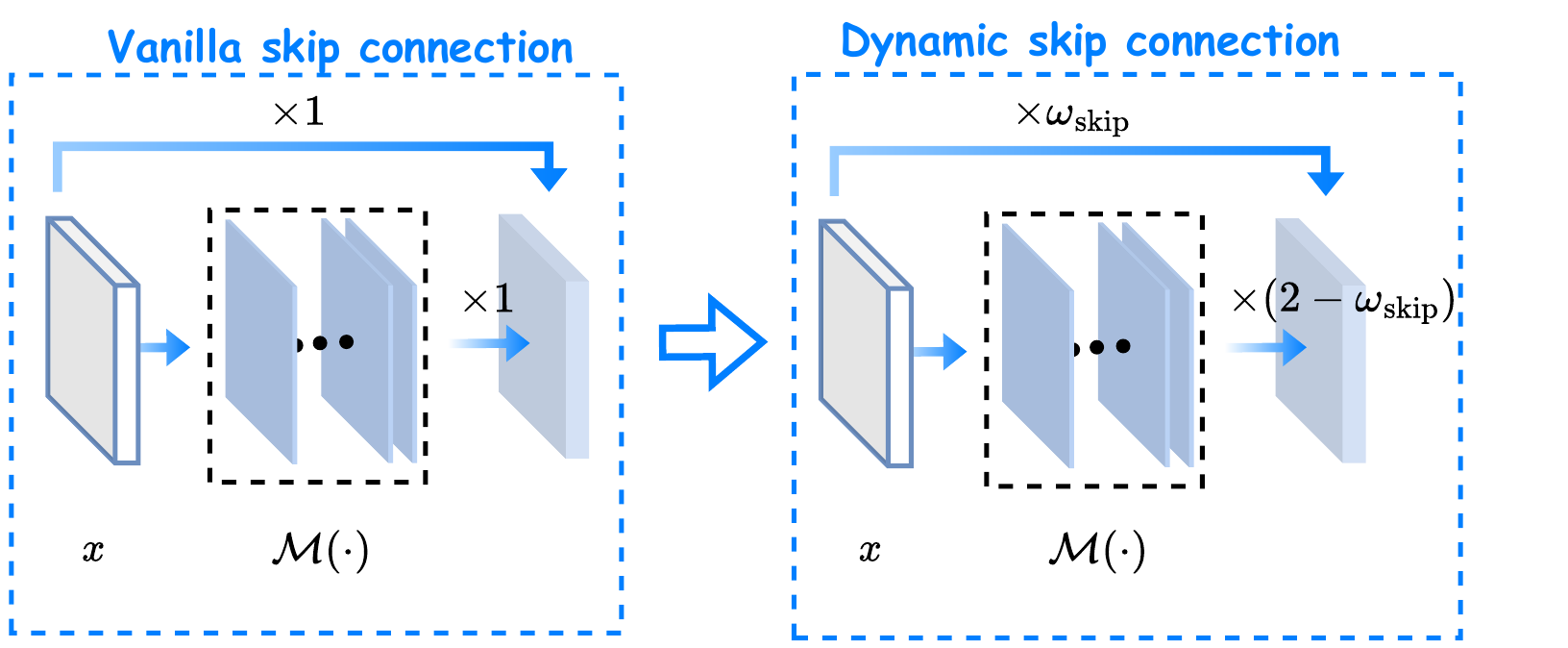}
\caption{The overview of vanilla skip connection and dynamic skip connection.}
\label{fig:dynamic_skip_connection}
\vspace{-15pt}
\end{wrapfigure} $t\rightarrow \epsilon$, the synthetic image's quality is superior, thus requiring a smaller step size to maintain stability.

\paragraph{Dynamic Skip Connection.} Training a velocity estimation model with a fixed architecture directly would be far from ideal. As demonstrated in Appendix~\ref{apd:cost_fo_fit_analysis}, the cost of fitting the training model varies at different $t$. In past work~\cite{nips22_design,song2019generative,2021pmlr_improved}, they both apply a special operation that skips the entire model with residual mapping, \textit{i.e.}, $c_\textrm{skip}(t)X_t+c_\textrm{out}(t)(X_t,t)$, where $c_\textrm{skip}(t)$ and $c_\textrm{out}(t)$ are two functions that input $t$ and output a scalar. However, this skip connection is coarse-grained and does not take advantage of the skip connection that contemporary mainstream neural networks~\cite{ResNet,ResNetv2,VIT,UNet} have themselves. A slight modification to the original UNet, \textit{i.e.}, implementing dynamic weights on the vanilla skip connection, allows for a reasonable allocation of the fitting overhead of the loss function at different $t$, and can improve DPM's performance simply and effectively. As shown in Fig.~\ref{fig:dynamic_skip_connection}, we implement dynamic skip connection by a simple linear dynamic weight $\omega_\textrm{skip}$. Thus, the novel dynamic skip connection can be denoted as
\renewcommand\arraystretch{1.1}
\setlength\tabcolsep{3pt}
\footnotesize
\begin{table}[t]
\center
\caption{{\bf Basic CUD:} only apply Runge-Kutta 12 and do not make use of strategies introduced in Sec.~\ref{sec:design_space}; {\bf RF:} Rectified flow; {\bf vanilla weight:} $\omega_1(i) \equiv 1$, $\omega_2(i) \equiv 1$; {\bf dynamic weight:} $\omega_1(i) = i/N$, $\omega_2(i) = 1-i/N$. The second column on the right represents the ODE solver and the number of sampling steps used. For example, ``Euler, 4'' represents the application of Euler's method with 4 steps for sampling. The results of IS and other loss functions are presented in Table~\ref{tab:ablation_study_result_1_appendix} and~\ref{tab:ablation_study_result_1_appendix_1} in Appendix.}
\label{tab:ablation_study_result_1}
\resizebox{1.\textwidth}{!}{%
\begin{tabular}{ll|r|r|r|r|r|r|r|r}\toprule
\multicolumn{2}{c|}{\multirow{1}{*}{Methods}} &  \multicolumn{1}{c|}{RF (Eq.~\ref{eq:curvature_ode})} & \multicolumn{3}{c|}{CD (train based on RF under continuous time steps scenarios)} & \multicolumn{4}{c}{Basic CUD} \\\hline
\multicolumn{2}{c|}{\multirow{2}{*}{Settings}} &\multicolumn{1}{c|}{ \multirow{2}{*}{MSE}} & \multicolumn{1}{c|}{\multirow{2}{*}{MSE, $h\rightarrow 0$}} & \multicolumn{1}{c|}{\multirow{2}{*}{MSE, $h\!=\!1/16$}}  & \multicolumn{1}{c|}{\multirow{2}{*}{LPIPS,$h\!=\!1/16$}} &   \multicolumn{1}{c|}{MSE,$h\!=\!1/16$,}  & \multicolumn{1}{c|}{LPIPS,$h\!=\!1/16$,} &  \multicolumn{1}{c|}{MSE,$h\!=\!1/16$,} &  \multicolumn{1}{c}{LPIPS,$h\!=\!1/16$,} \\
& & \multicolumn{1}{c|}{}& \multicolumn{1}{c|}{} & \multicolumn{1}{c|}{} & \multicolumn{1}{c|}{} & \multicolumn{1}{c|}{vanilla weight} & \multicolumn{1}{c|}{vanilla weight} & \multicolumn{1}{c|}{dynamic weight} & \multicolumn{1}{c}{dynamic weight} \\\hline
\multicolumn{1}{c|}{\multirow{4}{*}{FID}} & Euler, 4 & 37.05& 32.57 & \textcolor{C1}{417.36} & \textcolor{C1}{471.47} & 5.20 & \textcolor{C1}{446.45} &  \textcolor{C1}{171.88} &  \textcolor{C1}{438.21} \\
\multicolumn{1}{c|}{} & Euler, 16 & 6.85 & 6.36 &  \textcolor{C1}{413.11} &  \textcolor{C1}{476.98} & 13.96 & \textcolor{C1}{444.98} & \textcolor{C1}{188.99} & \textcolor{C1}{434.61} \\
\multicolumn{1}{c|}{} & Heun, 4 & 13.58 & 12.88 &  \textcolor{C1}{414.23} & \textcolor{C1}{477.17} & 6.28 & \textcolor{C1}{445.31} & \textcolor{C1}{195.27} & \textcolor{C1}{434.43} \\
\multicolumn{1}{c|}{} & Heun, 16 & 4.15 & 4.43 &  \textcolor{C1}{407.84} & \textcolor{C1}{473.55} & 17.62 & \textcolor{C1}{444.48} & \textcolor{C1}{193.47} & \textcolor{C1}{433.10} \\
\bottomrule
\end{tabular}}
\end{table}
\ablationifuseema
\ablationrandomstepshakedrop
\begin{equation}
\small
\begin{aligned}
& x_o = \omega_\textrm{skip}x + (2 - \omega_\textrm{skip})\mathcal{M}(x), \\
\end{aligned}
\label{eq:dynamic_skip_connection}
\end{equation}
where $x$, $x_o$, and $\mathcal{M}(\cdot)$ refer to the input, output, and intermediate layers, respectively. Furthermore, $\omega_\textrm{skip}=2(1-(t(1-2c_\textrm{skip})+c_\textrm{skip}))$ is an equation where $c_\textrm{skip}$ is a pre-set hyperparameter that we choose as either 0.25 or 0.75. Notably, we replace all vanilla skip connections in UNet with dynamic skip connections.
\section{Experiment}
We conduct experiments to evaluate the effectiveness of CUD on CIFAR-10~\cite{CIFAR} with a resolution of 32$\times$32, excluding the comparison experiments. We use three datasets for the comparison experiments: CIFAR-10 with resolution 32$\times$32, MNIST with resolution 28$\times$28, and ImageNet 64$\times$64 with resolution 64$\times$64. We assess the quality of the synthetic samples using Fr\'{e}chet Inception Distance (FID)~\cite{fid} and Inception Score (IS)~\cite{is}. To compute the FID, we compare 50,000 synthetic samples with all available real samples, and for computing the IS, we used 50,000 synthetic samples. We employed 4 different configurations for the velocity estimation model and hyperparameters on CIFAR-10, MNIST, and ImageNet 64$\times$64. Specifically, we used configurations (a), (b), and (c) to train DPMs on CIFAR-10, MNIST, and ImageNet 64$\times$64, respectively, and configuration (d) for our proposed final multi-step distillation (see Appendix~\ref{sec:final_multi_step_distillation}) on all datasets. The implementation details for these settings are provided in Appendix~\ref{apd:implementation_details}. Unless otherwise specified, configuration (a) was used for ablation studies and analyses.

\subsection{Ablation Study}
\paragraph{Basic CUD.} To verify that the basic CUD is capable of accelerated sampling, we conduct experiments and present the results in Table~\ref{tab:ablation_study_result_1}. In this table, the \textbf{\textit{dynamic weight}} represents the supervision focus during training. At the beginning of the training, \textbf{\textit{dynamic weight}} emphasizes using the ground truth label for supervision. In contrast, at the end of the training, \textbf{\textit{dynamic weight}} emphasizes using the output of the teacher model for supervision. And $L$ is one of Learned Perceptual Image Patch Similarity (LPIPS)~\cite{LPIPS} and MSE. LPIPS is more effective than MSE in Consistency Model~\cite{icml23_consistency}. But according to Table~\ref{tab:ablation_study_result_1}, we can conclude that MSE can work, but LPIPS does not, and that \textbf{\textit{vanilla weight}} can work but \textbf{\textit{dynamic weight}} does not. Meanwhile, the effective choice of loss function $L$, as well as the loss weights $\omega_1$ and $\omega_2$ in the basic CUD, are MSE, $\omega_1(i)\equiv 1$, and $\omega_2(i)\equiv 1$, respectively.

\paragraph{Investigating Design Space.} We present experimental results in Figs.~\ref{fig:ablation_study_if_use_ema} and~\ref{fig:ablation_study_if_use_random_step_and_shakedrop} to assess the efficacy of our proposed suitable strategies. In these figures, Runge-Kutta 12/23 represents the form utilized in Runge-Kutta-based multi-step alignment distillation, while $f_{\theta_-}$ and $f_{\theta}$ denote the employment of the EMA model and the training model for \textit{catch-up sampling}, respectively. Here, we do not consider the ground truth loss terms $\textrm{MSE}(\widetilde{X}_1-X_0,g_{\psi 2}(f_\theta(X_t,t)))$ and $\textrm{MSE}(\widetilde{X}_1-X_0,g_{\psi 3}(f_\theta(X_t,t)))$. In Appendix~\ref{apd:additional_experimental_results}, we perform ablation experiments to show their importance. For all intermediate sampling points, we employ velocity estimation using a one-step method to synthetic samples. Specifically, when obtaining an intermediate sampling point $Z_m$, we calculate the target ``clean'' image $Z_0$ using $Z_0 = Z_m - mf_\theta(Z_m,m)$ (Euler's method). From Fig.~\ref{fig:ablation_study_if_use_ema}, we deduce that employing the training model for \textit{catch-up sampling} is more beneficial than using the EMA model, suggesting the application of $f_\theta$ instead of $f_{\theta_-}$ in Eq.~\ref{eq:base_loss}. CUD is more effective in scenarios with fewer sampling steps but may be less effective than the baseline when more sampling steps are present. This is because CUD's primary purpose is to accelerate sampling rather than improve the quality of synthetic images. Consequently, the velocity estimation model should prioritize generating ``clean'' images in scenarios with fewer sampling steps rather than enhancing image quality in scenarios with more sampling steps. Additionally, in Fig.~\ref{fig:ablation_study_if_use_random_step_and_shakedrop}, we observe that both \textbf{\textit{uniform}} and \textbf{\textit{rule}} surpass the baseline in terms of synthetic image quality in nearly all scenarios, with greater improvements seen with fewer sampling steps. In fewer sampling steps scenarios, \textbf{\textit{rule}} outperforms \textbf{\textit{uniform}}, while the opposite is true for scenarios with more sampling steps (details are provided in Appendix~\ref{apd:additional_data_presentation}). Notably, \textbf{\textit{uniform}} achieves the best CUD performance, with an FID of 3.36. Lastly, the ablation experiments on dynamic skip connections are also presented in Figs.~\ref{fig:ablation_study_if_use_random_step_and_shakedrop}. Both $c_\textrm{skip}=$0.25 and $c_\textrm{skip}=$0.75 work very well because the fitting cost is minimized when $t$ is close to 0.5. In particular, $c_\textrm{skip}=$0.75 performs better than $c_\textrm{skip}=$0.25, achieving an FID 2.91 in 15 steps of sampling (see Appendix~\ref{apd:additional_data_presentation}). This suggests that to improve the quality of the synthetic images, the model should focus more on enhancing the representation at $t\rightarrow0$ rather than $t\rightarrow1$. By combining the above three suitable strategies that require no additional overhead and are simple yet effective, the best performance of CUD on FID has decreased from 5.20 (4 steps to sampling with Heun's method) to 2.91 (15 steps to sampling with Euler's method).

\paragraph{Runge-Kutta-Based Multi-Step Alignment Distillation.} We perform ablation studies to assess the efficacy of Runge-Kutta-based multi-step alignment distillation across different orders, specifically utilizing the CIFAR-10 and MNIST datasets. The corresponding results are displayed in Tables~\ref{tab:comparsion_results1} and~\ref{tab:comparsion_results2}. When examining the performance of Runge-Kutta orders 12, 23, and 34 on CIFAR-10, it becomes evident that images produced by Runge-Kutta 23 yield the most desirable performance. However, Runge-Kutta 34 attains the greatest accelerated sampling effect, followed by Runge-Kutta 23 and 12. In contrast, for the MNIST dataset, Runge-Kutta 34 provides the best performance and the most effective accelerated sampling. These empirical findings suggest that the accelerated sampling effect intensifies with the increase in the order of Runge-Kutta-based multi-step alignment distillation. Although the quality of images generated by Runge-Kutta 34 on CIFAR-10 is not as optimal as those produced by Runge-Kutta 12 and 23, it is justified due to the higher complexity of CIFAR-10 compared to MNIST, and the greater emphasis placed by Runge-Kutta 34 on improving accelerated sampling rather than enhancing the quality of the synthetic images.

\begin{table*}
    \begin{minipage}[t]{0.49\linewidth}
	\caption{Experimental results on CIFAR-10. $^\ast$Apply our proposed final multi-step distillation.}\label{tab:comparsion_results1}
	\centering
	{\setlength{\extrarowheight}{1.5pt}
	\begin{adjustbox}{max width=\linewidth}
	\begin{tabular}{lcccc}
        \Xhline{3\arrayrulewidth}
        \Xhline{3\arrayrulewidth}
	    METHOD & Solver & NFE ($\downarrow$) & FID ($\downarrow$) & IS ($\uparrow$) \\
           \\[-2.2ex]
       \multicolumn{4}{l}{\textbf{Generative Adversarial Network (GAN)}}\\\Xhline{3\arrayrulewidth}
       BigGAN~\cite{BigGAN} & - & 1 & 14.7 & 9.22 \\
       AutoGAN~\cite{AutoGAN} & - & 1 & 12.4 & 8.55 \\
       StyleGAN~\cite{StyleGAN} & - & 1 & 8.32 & 9.18 \\
       StyleGAN+ADA~\cite{StyleGAN} & - & 1 & 2.92 & 9.40 \\
        \\[-2ex]
        \multicolumn{4}{l}{\textbf{Diffusion Model (One Session)}}\\\Xhline{3\arrayrulewidth}
        DDPM~\cite{ddpm_begin} & - & 1000 & 3.21 & 9.46 \\
        \multirow{2}{*}{DDPM} & \multirow{2}{*}{\rowtl{DDIM}{\cite{ddim}}} & \multirow{2}{*}{10} & \multirow{2}{*}{8.23} & \multirow{2}{*}{-} \\
        \\
        \multirow{2}{*}{DDPM} & \multirow{2}{*}{\rowtl{DPM-solver-2}{\cite{dpm_solver}}} & \multirow{2}{*}{12} & \multirow{2}{*}{5.28} &  \multirow{2}{*}{-} \\
        \\
        DDPM & DPM-solver-3 & 12 & 6.03 & - \\
        NCSN++~\cite{sde} & Euler-Maruyama  & 2000 & \textbf{2.38} & 9.83 \\
        1-Rectified Flow$^+$~\cite{iclr22_rect} & Runge-Kutta 45 & 127 & 2.58 & 9.60 \\
        2-Rectified Flow$^+$ & Runge-Kutta 45 & 110 & 3.36 & 9.24 \\
        Curvature$^+$~\cite{icml23_curvature} & Euler & 16 &  6.85 & 8.84 \\
        Curvature$^+$ & Euler & 4 &  37.05 & 7.07 \\
        Curvature$^+$ & Heun & 7 &  13.58 & 8.49 \\   
        CUD (Runge-Kutta 12)$^+$ & Euler & 15 &  2.91 & \textbf{9.90} \\
        CUD (Runge-Kutta 12)$^+$ & Euler & 4 &  17.40 & 9.38 \\
        CUD (Runge-Kutta 12)$^+$ & DPM-solver-2 & 4 &  8.25 & 8.95 \\
        CUD (Runge-Kutta 23)$^+$ & Euler & 15 & 2.80 & 9.36 \\
        CUD (Runge-Kutta 23)$^+$ & Euler & 4 & 12.23 & 8.39 \\
        CUD (Runge-Kutta 23)$^+$ & DPM-solver-2 & 4 & 6.42 & 8.90 \\
        CUD (Runge-Kutta 34)$^+$ & Euler & 14 & 3.40 & 9.39 \\
        CUD (Runge-Kutta 34)$^+$ & Euler & 4 & 9.45 & 8.50 \\
        CUD (Runge-Kutta 34)$^+$ & Heun & 7 & 4.70 & 9.34 \\
            \\[-2.2ex]
        \multicolumn{4}{l}{\textbf{Diffusion Model (Two Session)}}\\\Xhline{3\arrayrulewidth}
        2-Rectified Flow+(distillation)$^+$ & Euler & 110 & 4.85 & 9.01 \\
        PD~\cite{iclr22_progressive} & DDIM & 1 & 8.34 & 8.69 \\
        PD & DDIM & 2 & 5.58 & 9.05 \\
        \multirow{2}{*}{CD~\cite{icml23_consistency}} & \multirow{2}{*}{\rowtl{Karra's method}{\cite{nips22_design}}} & \multirow{2}{*}{1} & \multirow{2}{*}{3.55} & \multirow{2}{*}{9.48} \\
        \\
        CD & Karra's method & 2 & 2.93 & 9.75 \\
        CUD (Runge-Kutta 12)+(distillation)$^*$$^+$ & Euler & 1 & 3.37 & 9.42 \\
        CUD (Runge-Kutta 23)+(distillation)$^*$$^+$ & Euler & 1  &  3.76 & 9.34 \\
        CUD (Runge-Kutta 34)+(distillation)$^*$$^+$ & Euler & 1 & 4.44& 9.09 \\\Xhline{3\arrayrulewidth}
	\end{tabular}
    \end{adjustbox}
	}
\end{minipage}
\hfill
\begin{minipage}[t]{0.49\linewidth}
	\caption{Experimental results on MNIST and ImageNet 64$\times$64. $^+$Train based on Rectified flow.}\label{tab:comparsion_results2}
	\centering
	{\setlength{\extrarowheight}{1.5pt}
	\begin{adjustbox}{max width=\linewidth}
	\begin{tabular}{lcccc}
	    \Xhline{3\arrayrulewidth}
        \Xhline{3\arrayrulewidth}
	    METHOD & Solver & NFE ($\downarrow$) & FID ($\downarrow$) & IS ($\uparrow$) \\
      \\[-2ex]
        \multicolumn{5}{l}{\textbf{MNIST (One Session)}}\\\Xhline{3\arrayrulewidth}
        1-Rectified flow$^+$~\cite{iclr22_rect} & Euler & 4 & 31.67 & 1.95 \\
        1-Rectified flow$^+$~\cite{iclr22_rect} & Euler & 16 & 11.16 & 2.06 \\
        1-Rectified flow$^+$~\cite{iclr22_rect} & Runge-Kutta 45 & 41 & 10.77 & 2.10 \\
        Curvature$^+$~\cite{icml23_curvature} & Euler & 16 & 21.77 & 2.11 \\
        Curvature$^+$ & Euler & 4 & 49.96 & 2.02 \\
        Curvature$^+$ & Heun & 7 & 21.55 & 2.10 \\
        CUD (Runge-Kutta 12)$^+$ & Euler & 16 & 3.39 & 2.01 \\
        CUD (Runge-Kutta 12)$^+$ & Euler & 4 & 9.45 & 2.02 \\
        CUD (Runge-Kutta 23)$^+$ & Euler & 16 & 2.18 & 2.10 \\
        CUD (Runge-Kutta 23)$^+$ & Euler & 4 & 7.06 & 2.03 \\
        CUD (Runge-Kutta 34)$^+$ & Euler & 16 & \textbf{1.81} & 2.10 \\
        CUD (Runge-Kutta 34)$^+$ & Euler & 4 & 6.70 & 2.03 \\
        \\[-2ex]
        \multicolumn{5}{l}{\textbf{MNIST (Two Session)}}\\\Xhline{3\arrayrulewidth}
        PD$^+$~\cite{iclr22_progressive} & Euler & 2 & 9.60 & 2.11 \\
        CD$^+$~\cite{icml23_consistency} & Euler & 2 & 8.96 & 2.05 \\
        CUD (Runge-Kutta 12)+(distillation)$^+$ & Euler & 1 & 8.94 & 2.08 \\
        CUD (Runge-Kutta 12)+(distillation)$^*$$^+$ & Euler & 1 & 6.38 & \textbf{2.12} \\
        CUD (Runge-Kutta 23)+(distillation)$^*$$^+$ & Euler & 1  & 3.08 & 2.08 \\
        CUD (Runge-Kutta 34)+(distillation)$^*$$^+$ & Euler & 1 & 5.43 & 2.09\\
        \Xhline{3\arrayrulewidth}
        \\
        \\
        \\
	    \Xhline{3\arrayrulewidth}
        \Xhline{3\arrayrulewidth}
	    METHOD & Solver & NFE ($\downarrow$) & FID ($\downarrow$) & Prec. ($\uparrow$) \\
        \\[-2ex]
        \multicolumn{5}{l}{\textbf{ImageNet 64$\times$64 (One Session)}}\\        \Xhline{3\arrayrulewidth}
        BigGAN-deep~\cite{BigGAN} & - & 1 & 4.06 & 0.79 \\
        IDDPM (small) & - & 4000 & 6.92 & 0.77 \\
        IDDPM (large) & - & 4000 & 2.92 & \textbf{0.82} \\
        ADM & DDIM & 250 & 2.61 & 0.73 \\
        ADM (dropout) & DDIM & 250 & \textbf{2.07} & 0.74 \\
        EDM & Karra's method & 79 & 2.44 & 0.71 \\
        U-ViT-M/4 & DPM-solver-2 & 50 & 5.85 & - \\
        U-ViT-L/4 & DPM-solver-2 & 50 & 4.26 & - \\
        CUD (Runge-Kutta 12)$^+$ & Euler & 4 & 18.84 &  0.58 \\
        CUD (Runge-Kutta 12)$^+$ & Euler & 8 & 7.75 & 0.68 \\
        CUD (Runge-Kutta 12)$^+$ & Euler & 16 & 5.05 & 0.71 \\
        CUD (Runge-Kutta 12)$^+$ & DPM-solver-2 & 6 & 7.68 & 0.68 \\
        \\[-2ex]
        \Xhline{3\arrayrulewidth}
	\end{tabular}
    \end{adjustbox}
    }
\end{minipage}
\vspace{-10pt}
\end{table*}
\subsection{Comparison Experiments}
The comparative experiments are performed on CIFAR-10, MNIST, and ImageNet 64$\times$64 to highlight the performance benefits of CUD. The results of PD~\cite{iclr22_progressive}, CD~\cite{icml23_consistency}, and Curvature~\cite{icml23_curvature} algorithms on MNIST are derived from Rectified flow, as the results of these methods are available for analysis. The relevant experimental results are presented in Tables~\ref{tab:comparsion_results1} and~\ref{tab:comparsion_results2}, and the remaining supplementary results are given in Appendix~\ref{apd:additional_experimental_results}. Considered as a one-session training approach, we first compare CUD with other one-session training methods, such as the train-free accelerated sampling techniques, namely ``DPM-solver'' and DDIM, as well as a variety of DPMs, specifically NCSN++, DDPM, Rectified Flow, and Curvature, which serves as the baseline. For experiments on all datasets, CUD is very effective in accelerated sampling. For instance, CUD achieves the best performance in the few-step image generation scenario. For instance, on CIFAR-10, a FID 4.70 obtained by a CUD (Runge-Kutta 34) sample in just 7 steps better than a FID 5.28 obtained by a DPM-solver-2 sample in 12 steps. 

Moreover, for the two-session training scenario, since CUD is trained on a continuous time step, it suffers from performance limitations. This is largely attributed to the excessive number of time points required for fitting, making it slightly inferior to both PD and CD algorithms. To address this, we introduced a novel multi-step distillation algorithm that extends the distillation algorithm presented in~\cite{iclr22_rect}, as a way to achieve one-step sampling. Specifically, the algorithm is designed to incrementally fit ``clean'' images obtained at different time steps (\textit{e.g.}, when t progresses from 1/2 to 5/16 to 1/8). The effectiveness of this approach has been validated through quantitative ablation experiments discussed in Appendix~\ref{sec:final_multi_step_distillation}. Leveraging this, we conduct distillation on the pre-trained model derived from CUD (Runge-Kutta 12), ultimately achieving a state-of-the-art FID of 3.37 by sampling in one step on CIFAR-10. Furthermore, our approach demonstrates cost-effectiveness in training, requiring only 620k (500k+120k) iterations on CIFAR-10 with a batch size of 128 and a model parameter number of 55M, compared to CD's 2100k (1300k+800k) iterations with a batch size of 256 and a model parameter number of 62M. Performance-wise, on both MNIST and ImageNet 64$\times$64, CUD demonstrates comparable results to PD and CD. Lastly, synthetic images generated by our approach can be found in Appendix~\ref{apd:synthetic_image}.
\section{Conclusion}
This paper presents Catch-Up Distillation (CUD), a method designed to integrate effortlessly with the existing diffusion model training paradigm, enabling high-quality image synthesis in fewer sampling steps, and eliminating the need for pre-training weights. The efficacy of CUD is substantiated through the validation of several datasets, including CIFAR-10, MNIST, and ImageNet 64$\times$64. In subsequent research, we sincerely hope that our CUD can be extended to discrete time steps where only a small number of time points need to be fitted, thus further enhancing the generality of CUD.

\clearpage
{\small
\bibliographystyle{abbrvnat}
\bibliography{egbib}

\begin{thebibliography}{47}
\providecommand{\natexlab}[1]{#1}
\providecommand{\url}[1]{\texttt{#1}}
\expandafter\ifx\csname urlstyle\endcsname\relax
  \providecommand{\doi}[1]{doi: #1}\else
  \providecommand{\doi}{doi: \begingroup \urlstyle{rm}\Url}\fi

\bibitem[Bao et~al.(2022)Bao, Li, Zhu, and Zhang]{iclr22_analytic}
F.~Bao, C.~Li, J.~Zhu, and B.~Zhang.
\newblock Analytic-dpm: an analytic estimate of the optimal reverse variance in
  diffusion probabilistic models.
\newblock In \emph{International Conference on Learning Representations},
  Virtual Event, Apr. 2022. OpenReview.net.

\bibitem[Brock et~al.(2019)Brock, Donahue, and Simonyan]{BigGAN}
A.~Brock, J.~Donahue, and K.~Simonyan.
\newblock Large scale gan training for high fidelity natural image synthesis.
\newblock In \emph{International Conference on Learning Representations}, New
  Orleans, LA, USA, May 2019. OpenReview.net.

\bibitem[Chuanguang et~al.(2021)Chuanguang, Zhulin, Linhang, and
  Yongjun]{HSAKD}
Y.~Chuanguang, A.~Zhulin, C.~Linhang, and X.~Yongjun.
\newblock Hierarchical self-supervised augmented knowledge distillation.
\newblock In \emph{Proceedings of the Thirtieth International Joint Conference
  on Artificial Intelligence}, pages 1217--1223, Virtual Event, Aug. 2021.
  IJCAI.

\bibitem[Dhariwal and Nichol(2021)]{iclr2021_guided_diffusion}
P.~Dhariwal and A.~Q. Nichol.
\newblock Diffusion models beat {GAN}s on image synthesis.
\newblock In \emph{Neural Information Processing Systems}, volume~34, pages
  8780--8794, Virtual Event, Dec. 2021. NIPS.
\newblock URL \url{https://openreview.net/forum?id=AAWuCvzaVt}.

\bibitem[Dosovitskiy et~al.(2020)Dosovitskiy, Beyer, Kolesnikov, Weissenborn,
  Zhai, Unterthiner, Dehghani, Minderer, Heigold, Gelly, et~al.]{VIT}
A.~Dosovitskiy, L.~Beyer, A.~Kolesnikov, D.~Weissenborn, X.~Zhai,
  T.~Unterthiner, M.~Dehghani, M.~Minderer, G.~Heigold, S.~Gelly, et~al.
\newblock An image is worth 16x16 words: Transformers for image recognition at
  scale.
\newblock In \emph{International Conference on Learning Representations}, Event
  Virtual, May 2020. OpenReview.net.

\bibitem[Du et~al.(2022)Du, Zhou, Feng, Tan, and Zhou]{nips2022_sam}
J.~Du, D.~Zhou, J.~Feng, V.~Tan, and J.~T. Zhou.
\newblock Sharpness-aware training for free.
\newblock In \emph{Advances in Neural Information Processing Systems},
  volume~35, pages 23439--23451, New Orleans, Louisiana, USA, Dec. 2022. NIPS.

\bibitem[Gong et~al.(2019)Gong, Chang, Jiang, and Wang]{AutoGAN}
X.~Gong, S.~Chang, Y.~Jiang, and Z.~Wang.
\newblock Autogan: Neural architecture search for generative adversarial
  networks.
\newblock In \emph{International Conference on Computer Vision}, pages
  3224--3234, Seoul, South Korea, Oct.-Nov. 2019. IEEE.

\bibitem[Goodfellow et~al.(2014)Goodfellow, Pouget-Abadie, Mirza, Xu,
  Warde-Farley, Ozair, Courville, and Bengio]{GAN_1}
I.~Goodfellow, J.~Pouget-Abadie, M.~Mirza, B.~Xu, D.~Warde-Farley, S.~Ozair,
  A.~Courville, and Y.~Bengio.
\newblock Generative adversarial nets.
\newblock In \emph{Neural Information Processing Systems}, volume~27, Long
  Beach, CA, USA, Jan. 2014. NIPS.

\bibitem[Gou et~al.(2021)Gou, Yu, Maybank, and Tao]{kdsurvey}
J.~Gou, B.~Yu, S.~J. Maybank, and D.~Tao.
\newblock Knowledge distillation: A survey.
\newblock \emph{International Journal of Computer Vision}, 129\penalty0
  (6):\penalty0 1789--1819, 2021.

\bibitem[He et~al.(2016{\natexlab{a}})He, Zhang, and Ren]{ResNet}
K.~He, X.~Zhang, and S.~Ren.
\newblock Deep residual learning for image recognition.
\newblock In \emph{Computer Vision and Pattern Recognition}, pages 770--778,
  Las Vegas, NV, USA, Jun. 2016{\natexlab{a}}. IEEE.

\bibitem[He et~al.(2016{\natexlab{b}})He, Zhang, Ren, and Sun]{ResNetv2}
K.~He, X.~Zhang, S.~Ren, and J.~Sun.
\newblock Identity mappings in deep residual networks.
\newblock In \emph{European Conference on Computer Vision}, pages 630--645,
  Amsterdam, North Holland, The Netherlands, Oct. 2016{\natexlab{b}}. Springer.

\bibitem[Heusel et~al.(2017)Heusel, Ramsauer, Unterthiner, Nessler, and
  Hochreiter]{fid}
M.~Heusel, H.~Ramsauer, T.~Unterthiner, B.~Nessler, and S.~Hochreiter.
\newblock Gans trained by a two time-scale update rule converge to a local nash
  equilibrium.
\newblock In \emph{Neural Information Processing Systems}, volume~30, Long
  Beach Convention Center, Long Beach, Dec. 2017. NIPS.

\bibitem[Hinton et~al.(2015)Hinton, Vinyals, and Dean]{vanillakd}
G.~Hinton, O.~Vinyals, and J.~Dean.
\newblock Distilling the knowledge in a neural network, 2015.
\newblock URL \url{https://arxiv.org/abs/1503.02531}.

\bibitem[Ho and Salimans(2021)]{nips2021_classifier_free_guidance}
J.~Ho and T.~Salimans.
\newblock Classifier-free diffusion guidance.
\newblock In \emph{Neural Information Processing Systems Workshop}, Virtual
  Event, Dec. 2021. NIPS.
\newblock URL \url{https://openreview.net/forum?id=qw8AKxfYbI}.

\bibitem[Ho et~al.(2020)Ho, Jain, and Abbeel]{ddpm_begin}
J.~Ho, A.~Jain, and P.~Abbeel.
\newblock Denoising diffusion probabilistic models.
\newblock In \emph{Neural Information Processing Systems}, pages 6840--6851,
  Virtual Event, Dec. 2020. NIPS.

\bibitem[Karras et~al.(2019)Karras, Laine, and Aila]{StyleGAN}
T.~Karras, S.~Laine, and T.~Aila.
\newblock A style-based generator architecture for generative adversarial
  networks.
\newblock In \emph{Computer Vision and Pattern Recognition}, pages 4401--4410,
  Seoul, South Korea, Oct.-Nov. 2019.

\bibitem[Karras et~al.(2022)Karras, Aittala, Aila, and Laine]{nips22_design}
T.~Karras, M.~Aittala, T.~Aila, and S.~Laine.
\newblock Elucidating the design space of diffusion-based generative models.
\newblock \emph{arXiv preprint arXiv:2206.00364}, 2022.

\bibitem[Kawar et~al.(2021)Kawar, Vaksman, and Elad]{nips21_b5c01503}
B.~Kawar, G.~Vaksman, and M.~Elad.
\newblock Snips: Solving noisy inverse problems stochastically.
\newblock In \emph{Neural Information Processing Systems}, volume~34, pages
  21757--21769, Virtual Event, Dec. 2021. NIPS.

\bibitem[Kingma et~al.(2021)Kingma, Salimans, Poole, and
  Ho]{nips2021_v_diffusion_model}
D.~Kingma, T.~Salimans, B.~Poole, and J.~Ho.
\newblock Variational diffusion models.
\newblock In \emph{Neural Information Processing Systems}, volume~34, pages
  21696--21707, Virtual Event, 2021. NIPS.

\bibitem[Kingma and Welling(2013)]{vae_3}
D.~P. Kingma and M.~Welling.
\newblock Auto-encoding variational bayes.
\newblock \emph{arXiv preprint arXiv:1312.6114}, 2013.

\bibitem[Kong et~al.(2021)Kong, Ping, Huang, Zhao, and Catanzaro]{audio_1}
Z.~Kong, W.~Ping, J.~Huang, K.~Zhao, and B.~Catanzaro.
\newblock Diffwave: A versatile diffusion model for audio synthesis.
\newblock In \emph{International Conference on Learning Representations},
  Virtual Event, May 2021. OpenReview.net.
\newblock URL \url{https://openreview.net/forum?id=a-xFK8Ymz5J}.

\bibitem[Krizhevsky et~al.(2009)Krizhevsky, Hinton, et~al.]{CIFAR}
A.~Krizhevsky, G.~Hinton, et~al.
\newblock Learning multiple layers of features from tiny images.
\newblock 2009.

\bibitem[Lee et~al.(2023)Lee, Kim, and Ye]{icml23_curvature}
S.~Lee, B.~Kim, and J.~C. Ye.
\newblock Minimizing trajectory curvature of ode-based generative models.
\newblock \emph{arXiv preprint arXiv:2301.12003}, 2023.

\bibitem[Li et~al.(2022)Li, Yang, Chang, Chen, Feng, Xu, Li, and
  Chen]{nc_sr_diffusion_model}
H.~Li, Y.~Yang, M.~Chang, S.~Chen, H.~Feng, Z.~Xu, Q.~Li, and Y.~Chen.
\newblock Srdiff: Single image super-resolution with diffusion probabilistic
  models.
\newblock \emph{Neurocomputing}, 479:\penalty0 47--59, 2022.
\newblock ISSN 0925-2312.
\newblock \doi{https://doi.org/10.1016/j.neucom.2022.01.029}.

\bibitem[Liu et~al.(2022)Liu, Gong, and Liu]{iclr22_rect}
X.~Liu, C.~Gong, and Q.~Liu.
\newblock Flow straight and fast: Learning to generate and transfer data with
  rectified flow.
\newblock \emph{arXiv preprint arXiv:2209.03003}, 2022.

\bibitem[Lu et~al.(2022{\natexlab{a}})Lu, Zhou, Bao, Chen, and
  Li]{dpm_solver++}
C.~Lu, Y.~Zhou, F.~Bao, J.~Chen, and C.~Li.
\newblock Dpm-solver++: Fast solver for guided sampling of diffusion
  probabilistic models.
\newblock \emph{arXiv preprint arXiv:2211.01095}, 2022{\natexlab{a}}.

\bibitem[Lu et~al.(2022{\natexlab{b}})Lu, Zhou, Bao, Chen, Li, and
  Zhu]{dpm_solver}
C.~Lu, Y.~Zhou, F.~Bao, J.~Chen, C.~Li, and J.~Zhu.
\newblock Dpm-solver: A fast ode solver for diffusion probabilistic model
  sampling in around 10 steps.
\newblock In \emph{Neural Information Processing Systems}, New Orleans, LA,
  USA, Nov.-Dec. 2022{\natexlab{b}}. NIPS.

\bibitem[Luhman(2021)]{arxiv21_kd_diffusion}
E.~Luhman.
\newblock Knowledge distillation in iterative generative models for improved
  sampling speed.
\newblock \emph{arXiv preprint arXiv:2101.02388}, 2021.

\bibitem[Meng et~al.(2022)Meng, Gao, Kingma, Ermon, Ho, and
  Salimans]{cvpr22_kd_guided}
C.~Meng, R.~Gao, D.~P. Kingma, S.~Ermon, J.~Ho, and T.~Salimans.
\newblock On distillation of guided diffusion models.
\newblock \emph{arXiv preprint arXiv:2210.03142}, 2022.

\bibitem[Mirzadeh et~al.(2020)Mirzadeh, Farajtabar, Li, Levine, Matsukawa, and
  Ghasemzadeh]{mirzadeh2020improved}
S.~I. Mirzadeh, M.~Farajtabar, A.~Li, N.~Levine, A.~Matsukawa, and
  H.~Ghasemzadeh.
\newblock Improved knowledge distillation via teacher assistant.
\newblock In \emph{Association for the Advance of Artificial Intelligence},
  volume~34, pages 5191--5198, New York, NY, USA, Feb. 2020. AAAI Press.

\bibitem[Nichol and Dhariwal(2021)]{2021pmlr_improved}
A.~Q. Nichol and P.~Dhariwal.
\newblock Improved denoising diffusion probabilistic models.
\newblock In \emph{International Conference on Machine Learning}, pages
  8162--8171. PMLR, 2021.

\bibitem[Park et~al.(2019)Park, Kim, Lu, and Cho]{RKD}
W.~Park, D.~Kim, Y.~Lu, and M.~Cho.
\newblock Relational knowledge distillation.
\newblock In \emph{Computer Vision and Pattern Recognition}, Long Beach, CA,
  USA, June 2019. IEEE.

\bibitem[Poole et~al.(2023)Poole, Jain, Barron, and
  Mildenhall]{iclr2023_dreamfusion}
B.~Poole, A.~Jain, J.~T. Barron, and B.~Mildenhall.
\newblock Dreamfusion: Text-to-3d using 2d diffusion.
\newblock In \emph{International Conference on Learning Representations}, 2023.
\newblock URL \url{https://openreview.net/forum?id=FjNys5c7VyY}.

\bibitem[Ronneberger et~al.(2015)Ronneberger, Fischer, and Brox]{UNet}
O.~Ronneberger, P.~Fischer, and T.~Brox.
\newblock U-net: Convolutional networks for biomedical image segmentation.
\newblock In \emph{Medical Image Computing and Computer-Assisted Intervention},
  pages 234--241, Germany, Central Europe, Oct. 2015. Springer.

\bibitem[Runge(1895)]{runge_kutta}
C.~Runge.
\newblock Ueber die numerische aufl\"{o}sung von differentialgleichungen.
\newblock pages 1432--1807, Jun. 1895.

\bibitem[Salimans and Ho(2022)]{iclr22_progressive}
T.~Salimans and J.~Ho.
\newblock Progressive distillation for fast sampling of diffusion models.
\newblock In \emph{International Conference on Learning Representations},
  Virtual Event, Apr. 2022. OpenReview.net.

\bibitem[Salimans et~al.(2016)Salimans, Goodfellow, Zaremba, Cheung, Radford,
  and Chen]{is}
T.~Salimans, I.~Goodfellow, W.~Zaremba, V.~Cheung, A.~Radford, and X.~Chen.
\newblock Improved techniques for training gans.
\newblock In \emph{Neural Information Processing Systems}, volume~29, Centre
  Convencions Internacional Barcelona, Barcelona SPAIN, Dec. 2016. NIPS.

\bibitem[Song et~al.(2021)Song, Meng, and Ermon]{ddim}
J.~Song, C.~Meng, and S.~Ermon.
\newblock Denoising diffusion implicit models.
\newblock In \emph{International Conference on Learning Representations},
  kigali, rwanda, May. 2021. OpenReview.net.

\bibitem[Song and Ermon(2019)]{song2019generative}
Y.~Song and S.~Ermon.
\newblock Generative modeling by estimating gradients of the data distribution.
\newblock In \emph{Neural Information Processing Systems}, volume~32. NIPS,
  2019.
\newblock URL
  \url{https://proceedings.neurips.cc/paper_files/paper/2019/file/3001ef257407d5a371a96dcd947c7d93-Paper.pdf}.

\bibitem[Song et~al.(2023{\natexlab{a}})Song, Dhariwal, Chen, and
  Sutskever]{icml23_consistency}
Y.~Song, P.~Dhariwal, M.~Chen, and I.~Sutskever.
\newblock Consistency models.
\newblock \emph{arXiv preprint arXiv:2303.01469}, 2023{\natexlab{a}}.

\bibitem[Song et~al.(2023{\natexlab{b}})Song, Sohl-Dickstein, Kingma, Kumar,
  Ermon, and Poole]{sde}
Y.~Song, J.~Sohl-Dickstein, D.~P. Kingma, A.~Kumar, S.~Ermon, and B.~Poole.
\newblock Score-based generative modeling through stochastic differential
  equations.
\newblock In \emph{International Conference on Learning Representations},
  kigali, rwanda, May. 2023{\natexlab{b}}. OpenReview.net.

\bibitem[Sun et~al.(2022)Sun, Chen, Wang, Ye, Feng, and
  Chen]{sun2022accelerating}
W.~Sun, D.~Chen, C.~Wang, D.~Ye, Y.~Feng, and C.~Chen.
\newblock Accelerating diffusion sampling with classifier-based feature
  distillation.
\newblock \emph{arXiv preprint arXiv:2211.12039}, 2022.

\bibitem[Vahdat and Kautz(2020)]{vae_2}
A.~Vahdat and J.~Kautz.
\newblock Nvae: A deep hierarchical variational autoencoder.
\newblock volume~33, pages 19667--19679, Virtual Event, Dec. 2020. NIPS.

\bibitem[Yang et~al.(2022)Yang, Zhang, Song, Hong, Xu, Zhao, Shao, Zhang, Cui,
  and Yang]{yang2022diffusion}
L.~Yang, Z.~Zhang, Y.~Song, S.~Hong, R.~Xu, Y.~Zhao, Y.~Shao, W.~Zhang, B.~Cui,
  and M.-H. Yang.
\newblock Diffusion models: A comprehensive survey of methods and applications.
\newblock \emph{arXiv preprint arXiv:2209.00796}, 2022.

\bibitem[Zhang and Chen(2023)]{zhang2023fast}
Q.~Zhang and Y.~Chen.
\newblock Fast sampling of diffusion models with exponential integrator.
\newblock In \emph{International Conference on Learning Representations}.
  OpenReview.net, 2023.
\newblock URL \url{https://openreview.net/forum?id=Loek7hfb46P}.

\bibitem[Zhang et~al.(2018)Zhang, Isola, Efros, Shechtman, and Wang]{LPIPS}
R.~Zhang, P.~Isola, A.~A. Efros, E.~Shechtman, and O.~Wang.
\newblock The unreasonable effectiveness of deep features as a perceptual
  metric.
\newblock In \emph{Computer Vision and Pattern Recognition}, Salt Lake City,
  Utah, USA, June 2018. IEEE.

\bibitem[Zhu et~al.(2017)Zhu, Park, Isola, and Efros]{GAN_2}
J.-Y. Zhu, T.~Park, P.~Isola, and A.~A. Efros.
\newblock Unpaired image-to-image translation using cycle-consistent
  adversarial networks.
\newblock In \emph{International Conference on Computer Vision}, pages
  2223--2232. IEEE, 2017.

\end{thebibliography}
}

\clearpage
\appendix
\section{Transport Cost under Reparameterized Noise Encoder $p_\psi(\cdot)$}
\label{apd:transport_cost}
For a coupling $(X_0,X_1)$, where $X_0\sim \pi_0,\ X_1\sim \pi_1$ and satisfies $p(\pi_0,\pi_1) = p(\pi_0)p(\pi_1)$, we have the Mutual Information (MI) between $\pi_0$ and $\pi_1$ is $0$, \textit{a.k.a.}, $I(\pi_0,\pi_1)=0$. Following~\cite{icml23_curvature}, let us define a reparameterized noise encoder $q_\psi(\cdot)$ that satisfies $\widetilde{X}_1 = q_\psi(X_0)$ and $\min \textrm{D}_{KL}(q_\psi(\pi_1|\pi_0)||p(\pi_1))$. When the cost function $\mathbb{E}[c(X_0-X_1)]\propto 1/I(\textrm{Law}(X_0),\textrm{Law}(X_1))$, we have $\mathbb{E}[c(X_0-\widetilde{X}_1)] \leq \mathbb{E}[c(X_0-X_1)]$.
\begin{proof}
\begin{equation}
\small
\begin{aligned}
\mathbb{E}[c(X_0-\widetilde{X}_1)] &  = k/ I(\textrm{Law}(X_0),\textrm{Law}(\widetilde{X}_1))  \\
& = k/\textrm{D}_{KL}(p(\pi_0,\pi_1)||p(\pi_0)q_\psi(\pi_1|\pi_0)) \\
& \leq  k/\textrm{D}_{KL}(p(\pi_0,\pi_1)||p(\pi_0)p(\pi_1))  \\
& =  k/I(\textrm{Law}(X_0),\textrm{Law}({X}_1)) \\
& =\mathbb{E}[c(X_0-{X}_1)], \\
\end{aligned}
\label{eq:mi_cost_function}
\end{equation}
where $k\in \mathbb{R}^+$ denotes a constant.
\end{proof}

\section{Consistency Distillation and Consistency Training}
\label{apd:cdt}
Pseudo-code and implementation details of Consistency Distillation (CD) and Consistency Training (CT) can be found in paper \textcolor[rgb]{1,0.078,0.58}{\url{https://arxiv.org/abs/2303.01469}} and github link \textcolor[rgb]{1,0.078,0.58}{\url{https://github.com/openai/consistency_models}}. Although Song \textit{et. al.} state that CD is capable of implementing the new state-of-the-art FID 3.55 under the condition of a single NFE, there still are some constraints on this:
\paragraph{Constraint I.} CD and CT may not be effective for certain popular DPMs such as Rectified flow~\cite{iclr22_rect}, NCSN++, and DDPM~\cite{sde}. When CD or CT is applied to Rectified flow under the continuous time steps scenarios, it causes the training to collapse in our experiments (Table~\ref{tab:ablation_study_result_1}). This means that CD/CT requires Karra's method under discrete time steps scenarios for it to work properly.
\paragraph{Constraint II.} CT can only be applied in \textit{empirical PF ODE} because if the model design for estimating ``clean image'', \textit{i.e.}, $c_\textrm{skip}(\cdot),c_\textrm{out}(\cdot)$, in the DPM is ignored, then it is equivalent to its not having the ground truth label for effective supervision. For CT in \textit{empirical PF ODE}, it relies on $f_\theta(X_t,t) = c_\textrm{skip}(t)X_t+c_\textrm{out}(t)F_\theta(X_t,t)$ to achieve implicit supervision. To be specific, CT's core loss function $||f_\theta(X_{t+h},t+h)-f_\theta(X_{t},t)||_2^2$ can be rewritten as $||c_\textrm{skip}(t+h)X_{t+h}-c_\textrm{skip}(t)X_{t}+c_\textrm{out}(t+h)F_\theta(X_{t+h},t+h)-c_\textrm{out}(t)F_\theta(X_{t},t)||_2^2$. Setting $X_t = X_0 + tX_1,\ X_1\sim \mathcal{N}(0,\mathbf{I})$, $c_\textrm{skip}(t) = 1-t$ and $c_\textrm{out}(t) = t$, the loss function can continue to be rewritten as $||t(F_\theta(X_{t+h},t+h)-F_\theta(X_{t},t)) + hF_\theta(X_{t+h},t+h) -h(X_0-(1-2t)X_1)-h^2X_1||_2^2$. Since \textit{empirical PF ODE} guarantees that $f_\theta(X_t,t)$ is an identity function when $t=0$\footnote{The original paper is that $f_\theta(X_t,t) = X_t$ when $t = \epsilon$. However, $c_\textrm{skip}(t)$ from the same study is defined as $\frac{1}{1+2(t-\epsilon)^2}$. Therefore, our derivations align theoretically in both instances.}, \textit{i.e.}, $f_\theta(X_0,0) = X_0$, $F_\theta(X_{t+h},t+h)$ is applied to fit the ground truth label $\frac{tF_\theta(X_t,t)+hX_0+h(2t-1)X_1+h^2X_1}{t+h}$. If $t=0$, $f_\theta(X_h,h)$ is required to fit $X_0+(h-1)X_1$. Then, if $t=h$, $f_\theta(X_{2h},2h)$ is required to fit $X_0+(2h-1)X_1$... Therefore, through a chain reaction, if $t=ih$, $f_\theta(X_{(i+1)h},(i+1)h)$ is required to fit $X_0+((i+1)h-1)X_1$.

Of course, this exquisite form of chain constraint accomplishes effective supervision, but it also hinders its possible application to other DPMs.

\paragraph{Constraint III.} Although the CT algorithm is able to accomplish distillation without relying on pre-trained weights for accelerated sampling, its application limitations and not very impressive performance prevent it from being a general-purpose algorithm. Our proposed CUD exists to address this problem, and it is theoretically applicable to any continuous SDE/ODE algorithm \textit{w.r.t.}, for SDE Runge-Kutta-based multi-step alignment distillation can be done with DDIM~\cite{ddim}.

\paragraph{Constraint IV.} The models utilized in CD/CT, possessing 62 million parameters, are larger than our proposed CUD, which encompasses 55 million parameters in the CIFAR-10 dataset. CD requires 80w iterations for training, not including the additional 130w iterations necessary to train the original diffusion model. In contrast, CUD, despite incorporating multi-step distillation, only necessitates a total of 62w iterations, broken down into 50w and 12w iterations, respectively.

\section{The Working Principle of Base Loss $\mathcal{L}_\textrm{base}$}
\label{apd:the_working_principle}
Before explaining why our proposed base loss can work, we first need to explain why Consistency Distillation (CD) of Consistency Model cannot work. As you know, the loss function of CD can be denoted as 
$$\mathcal{L}_\textrm{CT} = L(f_\theta(X_t,t),f_{\theta_-}(X_t-hf_{\Psi}(X_t,t),t-h)),$$
where $L(\cdot,\cdot)\in \mathbb{R}^+$, $\theta_-$, and $\Psi$ denote a distance function, the parameters of the EMA model, and the parameters of the pre-trained model, respectively. Assume that $\operatorname*{sup}_{X_t} L(f_\theta(X_t,t),f_{\theta_-}(X_t-hf_{\Psi}(X_t,t),t-h))\leq \gamma_1$ and $\operatorname*{sup}_{X_t} L(f_\theta(X_t,t),f_{\theta_-}(X_t,t))\leq \gamma_2$ after training convergence. And $f_\theta(\cdot,t)$ satisfies Lipschitz condition, \textit{i.e.}, $||f_\theta(x,t)-f_\theta(y,t)|| \leq K ||x-y||$, there exists $K>0$ such that $\forall$ $t \in [\epsilon,1]$. We can expand $\operatorname*{sup}_{X_t} L(f_\theta(X_t,t),f_{\theta_-}(X_t-hf_{\Psi}(X_t,t),t-h))\leq \gamma_1$ as
\begin{equation}
\small
\begin{aligned}
& \operatorname*{sup}_{X_t} L(f_\theta(X_t,t),f_{\theta_-}(X_t-hf_{\Psi}(X_t,t),t-h))\leq \gamma_1 \\
=>& \operatorname*{sup}_{X_t} L(f_\theta(X_t,t),f_{\theta}(X_t-hf_{\Psi}(X_t,t),t-h))-\operatorname*{sup}_{X_t} L(f_{\theta}(X_t-hf_{\Psi}(X_t,t),t-h),f_{\theta_-}(X_t-hf_{\Psi}(X_t,t),t-h))\leq \gamma_1 \\ 
=>& \operatorname*{sup}_{X_t} L(f_\theta(X_t,t),f_{\theta}(X_t-hf_{\Psi}(X_t,t),t-h))\leq \gamma_1 + \gamma_2 \\ 
=>&  L(f_\theta(X_t,t),f_\theta(X_{t-h},t-h))\leq L(f_\theta(X_{t-h},t-h),f_\theta(X_t-hf_{\Psi}(X_t,t),t-h)) +\gamma_1 + \gamma_2.\\
\end{aligned}
\label{eq:proof_our_core_function_1}
\end{equation}
Suppose that $L(\cdot,\cdot)$ is a norm, \textit{i.e.}, $||\cdot-\cdot||$. Then we can continue to derive Eq.~\ref{eq:proof_our_core_function_1} as
\begin{equation}
\small
\begin{aligned}
&||f_\theta(X_t,t)-f_\theta(X_{t-h},t-h)||\leq  Kh||X_1-X_0,f_\Psi(X_t,t)|| +\gamma_1 + \gamma_2.\\
\end{aligned}
\label{eq:proof_our_core_function_2}
\end{equation}
$\forall t_a,t_b\in [\epsilon,1]$ satisfy $\epsilon \leq t_a \leq t_b \leq 1$, we have
\begin{equation}
\small
\begin{aligned}
& ||f_\theta(X_{t_b},t_b) - f_\theta(X_{t_a},t_a)|| \leq  \frac{t_b-t_a}{h}\left[\gamma_1+\gamma_2+Kh||f_\Psi(X_{t_b},t_b)-(X_1-X_0)||\right].\\
\end{aligned}
\label{eq:proof_our_core_function_3}
\end{equation}
The right half of the above equation contains three terms: $\gamma_1$, $\gamma_2$, and $Kh||f_\Psi(X_{t_b},t_b)-(X_1-X_0)||$. Only these terms are small enough to ensure the stability of the distillation. CD ignores the difference $\gamma_2$ between the EMA model and the training model, which causes the training to collapse extremely easily under continuous time steps scenarios, especially if the EMA model is not updated accurately at one step.

If changing the pre-trained weight $\Phi$ to $\theta$, the left-hand side of Eq.~\ref{eq:proof_our_core_function_3} can be used directly to complete the distillation process. This form is different from the standard DPM training and is already widely applied in other applications to a certain extent. In image classification tasks, distillation algorithms typically involve two core losses: (1) Cross-Entropy loss between the student model output and the ground truth label, and (2) Kullback-Leibler Divergence between the student model output and the teacher model output. Collaborative supervision between the teacher model output and the ground truth label significantly enhances the generalization ability of the student model. This approach can also be applied to the DPM for distillation without teacher weights, thus achieving better performance of the student model.

Therefore, a distance function in the form of a norm as a loss function and the application of the ground truth label for supervision is necessary for the DPM. Only by ensuring these two points, Eq.~\ref{eq:proof_our_core_function_3}'s bound does not collapse into training by being too lenient.
\CUDtwo
\CUDthree
\section{Derivation of Runge-Kutta-Based Multi-Step Alignment Distillation}
\label{apd:derivation_rk_msad}
For the ODE $\frac{dX_t}{dt} = g_{\psi 1}(f_\theta(X_t,t))$, the precision of estimating $\widetilde{X}_{t-h}$ is crucial for performing \textit{catch-up sampling} from $X_t$. Specifically, the smaller the truncation error of the ODE solver, the more precise the numerical integration will be in obtaining $X_{t-h}$. Runge-Kutta methods, which include Euler's method and Heun's methods, enable a consistent view of modeling \textit{catch-up sampling}. In particular, Euler's method is a first-order ODE solver with truncation error $\mathcal{O}(h^2)$, and Heun's method is a second-order ODE solver with truncation error $\mathcal{O}(h^3)$. Euler's method is the same as Runge-Kutta 12 but Heun's method is a subset of Runge-Kutta 23, due to the number of constraint equations of Runge-Kutta 23 being less than the number of solution factors. We only consider Runge-Kutta 12, Runge-Kutta 23, and Runge-Kutta 34 in this work because of the additional computational overhead required for higher orders Runge-Kutta methods. Let us define \textit{catch-up sampling} as $\widetilde{X}_{t-h} = X_t - \Phi(X_t,g_{\psi 1}(f_\theta(\cdot,\cdot)),h)$, where $\Phi$ is the update function of a one-step ODE solver that takes $X_t$, $g_{\psi 1}(f_\theta(\cdot,\cdot))$ and $h$ as inputs to estimate the integration of velocities $-\int_{t}^{t-h}g_{\psi 1}(f_\theta(\widetilde{X}_\tau,\tau))d\tau$. Based on this, Runge-Kutta methods can be modeled out as
\begin{equation}
\small
\begin{aligned}
& k_1 = hg_{\psi 1}(f_\theta(X_t,t)),\quad\\
& k_2 = hg_{\psi 1}(f_\theta(X_t-\sum_{j=1}^1b_{2j}k_j,t-a_2h)),\quad\\
& k_3 = hg_{\psi 1}(f_\theta(X_t-\sum_{j=1}^2b_{3j}k_j,t-a_3h)),\quad\quad\quad\quad\quad \\
\end{aligned}
\label{eq:runge_kutta_define_1}
\end{equation}
\begin{equation}
\small
\begin{aligned}
& \cdots  \\
& k_i = hg_{\psi 1}(f_\theta(X_t-\sum_{j=1}^{i-1}b_{ij}k_j,t-a_ih)),\\
& \widetilde{X}_{t-h} = X_t - \Phi(X_t,g_{\psi 1}(f_\theta(\cdot,\cdot)),h) \approx X_t - \sum_{j=1}^i\omega_jk_j,\\
\end{aligned}
\label{eq:runge_kutta_define}
\end{equation}
where $i$ refers to the order of the Runge-Kutta method and ensures $i>1$. Meanwhile, the correlation factors in set $\mathbb{W} := \{b_{2j}\}_{j=1}^1\cup \{b_{3j}\}_{j=1}^2 \cup\cdots\cup \{b_{ij}\}_{j=1}^{i-1}\cup \{a_j\}_{j=2}^{i} \cup \{\omega\}_{j=1}^i$ are unknown. If $i=1$, then all coefficients will not exist and the Runge-Kutta method reduces to Euler's method, \textit{a.k.a}, Runge-Kutta 12.

The factors in $\mathbb{W}$ without any constraints cannot produce Runge-Kutta 23 and Runge-Kutta 34. More specifically, if the order of the corresponding Runge-Kutta method is $i$, then its truncation error needs to be under certain constraints (\textit{i.e.}, align these factors by Taylor expansion) to reach $\mathcal{O}(h^{i+1})$. Based on the derivation in~\cite{runge_kutta}, we can conclude that the conditions for achieving Runge-Kutta 23 and Runge-Kutta 34 are needed to satisfy
\begin{equation}
\small
\begin{aligned}
\textcolor{C3}{\#\textrm{Runge-Kutta 23:}} &   \\
&  \omega_1+\omega_2 = 1,\ a_2\omega_2 = \frac{1}{2} \\
&  b_{21}\omega_2 = \frac{1}{2},\ a_2 = b_{21} \\
\textcolor{C3}{\#\textrm{Runge-Kutta 34:}} &   \\
&  \omega_1+\omega_2+\omega_3 = 1,\ a_{2} = b_{21}, \\
& a_{3} = b_{31} + b_{32},\ \omega_2a_2+\omega_3a_3 = \frac{1}{2}, \\
& \omega_2a_2^2+\omega_3a_3^2 = \frac{1}{3},\ \omega_2b_{32}a_2 = \frac{1}{6}, \\
\end{aligned}
\label{eq:runge_kutta_derivation}
\end{equation}
As shown in Fig.~\ref{fig:cud_overall_structure}, our objective is to sample several points and utilize Runge-Kutta methods to compute the solutions of ODE at multiple instances starting from these points. Subsequently, we aim to estimate the velocities from the obtained solutions and align them with the outputs of the various heads $\{g_{\psi j}(\cdot)\}_{j=1}^i$. This means that we need to get the estimate of $\widetilde{X}_{t-h},\widetilde{X}_{t-2h},\widetilde{X}_{t-3h}$ and expand them by Taylor's Theorem in $t$, then the formula can be written as
\begin{equation}
\small
\begin{aligned}
& \widetilde{X}_{t-jh} = X_t - (jh)\frac{d X_t}{dt} + (jh)^2 \frac{d^2 X_t}{dt^2} - (jh)^3  \frac{d^3 X_t}{dt^3} +  (jh)^4  \frac{d^4 X_t}{dt^4} + P_4(t-jh),\ s.t.,\ j \in \{1,2,3\} \\
\end{aligned}
\label{eq:runge_kutta_taylor}
\end{equation}
where $P_4(t-jh)$ represents the remainder term. These differential terms $\{\frac{d^iX_t}{dt^i}\}_{i=1}^4$ are also the derivatives of different orders of the function $g_{\psi 1}(f_{\theta}(\cdot,\cdot))$. The very crucial point is that in the Taylor expansion, our pre-set step $h$ changes to $jh$, and this scaling can in fact be transferred to the factors in $\mathbb{W}$. Thus, Eq.~\ref{eq:runge_kutta_derivation} can be rewritten as
\begin{equation}
\small
\begin{aligned}
\textcolor{C3}{\#\textrm{Runge-Kutta 23:}} &   \\
&  \omega_1+\omega_2 = j,\ a_2\omega_2 = \frac{j}{2} \\
&  b_{21}\omega_2 = \frac{j}{2},\ a_2 = b_{21} \\
\textcolor{C3}{\#\textrm{Runge-Kutta 34:}} &   \\
&  \omega_1+\omega_2+\omega_3 = j,\ a_{2} = b_{21}, \\
& a_{3} = b_{31} + b_{32},\ \omega_2a_2+\omega_3a_3 = \frac{j}{2}, \\
& \omega_2a_2^2+\omega_3a_3^2 = \frac{j}{3},\ \omega_2b_{32}a_2 = \frac{j}{6}. \\
\end{aligned}
\label{eq:runge_kutta_derivation_new}
\end{equation}
This rewrite is the result of transferring the effect of $h\rightarrow jh$ to $\{\omega_j\}_{j=1}^i$. In Eq.~\ref{eq:runge_kutta_derivation_new}, we have a necessary constraint that $\forall i\in \{1,2,3\}$, the factors in $\{b_{2j}\}_{j=1}^1\cup \{b_{3j}\}_{j=1}^2 \cup\cdots\cup \{b_{ij}\}_{j=1}^{i-1}\cup \{a_j\}_{j=2}^{i}$ must remain fixed, since we need to ensure that all sampling points remain fixed. Otherwise, the computational overhead would increase exponentially. Then, we can obtain the following novel constraints:
\begin{equation}
\small
\begin{aligned}
\textcolor{C3}{\#\textrm{Runge-Kutta 23:}} &   \\
&   a_2 =b_{21}, \\
\textcolor{C3}{\#\textrm{Runge-Kutta 34:}} &   \\
&   a_3 - a_2 = (3a_3 - 3a_2+1) b_{32}.\\
\end{aligned}
\label{eq:runge_kutta_derivation_new_2}
\end{equation}
This means that we only need to satisfy the above constraint, and then multi-step alignment distillation based on Runge-Kutta 23 and Runge-Kutta 34 can be achieved. A natural form of sampling is equidistant sampling, \textit{i.e.}, $a_3 = 2,a_2=1$. Based on this, all the factors can be determined:
\begin{equation}
\small
\begin{aligned}
\textcolor{C3}{\#\textrm{Runge-Kutta 23:}} &   \\
&   a_2 =1, b_{21}=1, \\
& \omega_1 = \frac{j}{2}, \omega_{2} = \frac{j}{2}, \\
\textcolor{C3}{\#\textrm{Runge-Kutta 34:}} &   \\
&   a_2 = 1, a_3 = 2, b_{21} = 1, b_{31} = \frac{7}{4} , b_{32} = \frac{1}{4},\\
& \omega_{1}=\frac{5j}{12},\omega_2 = \frac{2j}{3},\omega_{3} =  -\frac{j}{12}. \\
\end{aligned}
\label{eq:runge_kutta_derivation_new_3}
\end{equation}
We can derive the following Runge-Kutta-based multi-step alignment distillation algorithm:
\begin{equation}
\small
\begin{aligned}
\textcolor{C3}{\#\textrm{Runge-Kutta 23:}} &   \\
& k_1=g_{\psi 1}(f_\theta(X_t,t)),k_2=g_{\psi 1}(f_\theta(X_t-hk_1,t-h)), \\
& \widetilde{X}_{t-h} = X_t - [\frac{1}{2}hk_1+\frac{1}{2}hk_2] + \mathcal{O}(h^3), \\
&  \widetilde{X}_{t-2h} = X_t - [hk_1+hk_2] +  \mathcal{O}(8h^3), \\
& v_{t-h} = g_{\psi 1}(f_\theta(\widetilde{X}_{t-h},t-h)), v_{t-2h} = g_{\psi 1}(f_\theta(\widetilde{X}_{t-2h},t-2h)), \\
&\mathcal{L}_{kd} = \sum_{j=1}^2\textrm{MSE}(v_{t-jh},g_{\psi j}(f_\theta(X_t,t))). \\
\textcolor{C3}{\#\textrm{Runge-Kutta 34:}} &   \\
& k_1=g_{\psi 1}(f_\theta(X_t,t)),k_2=g_{\psi 1}(f_\theta(X_t-hk_1,t-h)),k_3=g_{\psi 1}(f_\theta(X_t-\frac{7}{4}hk_1-\frac{1}{4}hk_2,t-2h)), \\
& \widetilde{X}_{t-h} = X_t - [\frac{5}{12}hk_1+\frac{2}{3}hk_2-\frac{1}{12}hk_3] + \mathcal{O}(h^4), \\
&  \widetilde{X}_{t-2h} = X_t - [\frac{5}{6}hk_1+\frac{4}{3}hk_2-\frac{1}{6}hk_3] + \mathcal{O}(16h^4), \\
&  \widetilde{X}_{t-3h} = X_t - [\frac{5}{4}hk_1+2hk_2-\frac{1}{4}hk_3] + \mathcal{O}(81h^4), \\
& v_{t-h} = g_{\psi 1}(f_\theta(\widetilde{X}_{t-h},t-h)), v_{t-2h} = g_{\psi 1}(f_\theta(\widetilde{X}_{t-2h},t-2h)), v_{t-3h} = g_{\psi 1}(f_\theta(\widetilde{X}_{t-3h},t-3h)), \\
&\mathcal{L}_{kd} = \sum_{j=1}^3\textrm{MSE}(v_{t-jh},g_{\psi j}(f_\theta(X_t,t))). \\
\end{aligned}
\label{eq:runge_kutta_derivation_new_4}
\end{equation}
% It should be noted that additional weights, $\frac{1}{j^3}$ or $\frac{1}{j^4}$, are incorporated in front of the loss functions to mitigate the adverse effects of varying estimation errors produced by the Runge-Kutta method at distinct time steps. The weighting of the corresponding loss function is determined empirically, derived from the reciprocal form of the truncation error, while disregarding the factor $h$.
\section{Higher-Order Runge-Kutta-Based Multi-step Alignment Distillation Algorithm}
\label{sec:rkb_ms_ada}
Due to space limitations in the main paper, we present here procedures of the higher order Runge-Kutta-based multi-step alignment distillation approaches, \textit{i.e.}, Runge-Kutta 23 (Algorithm~\ref{alg:CUDtwo}) and Runge-Kutta 34 (Algorithm~\ref{alg:CUDthree}).

\begin{wrapfigure}{r}{3.2cm}
\vspace{-50pt}
\includegraphics[height=0.2\textwidth,trim={0cm 0cm 1.8cm 2cm},clip]{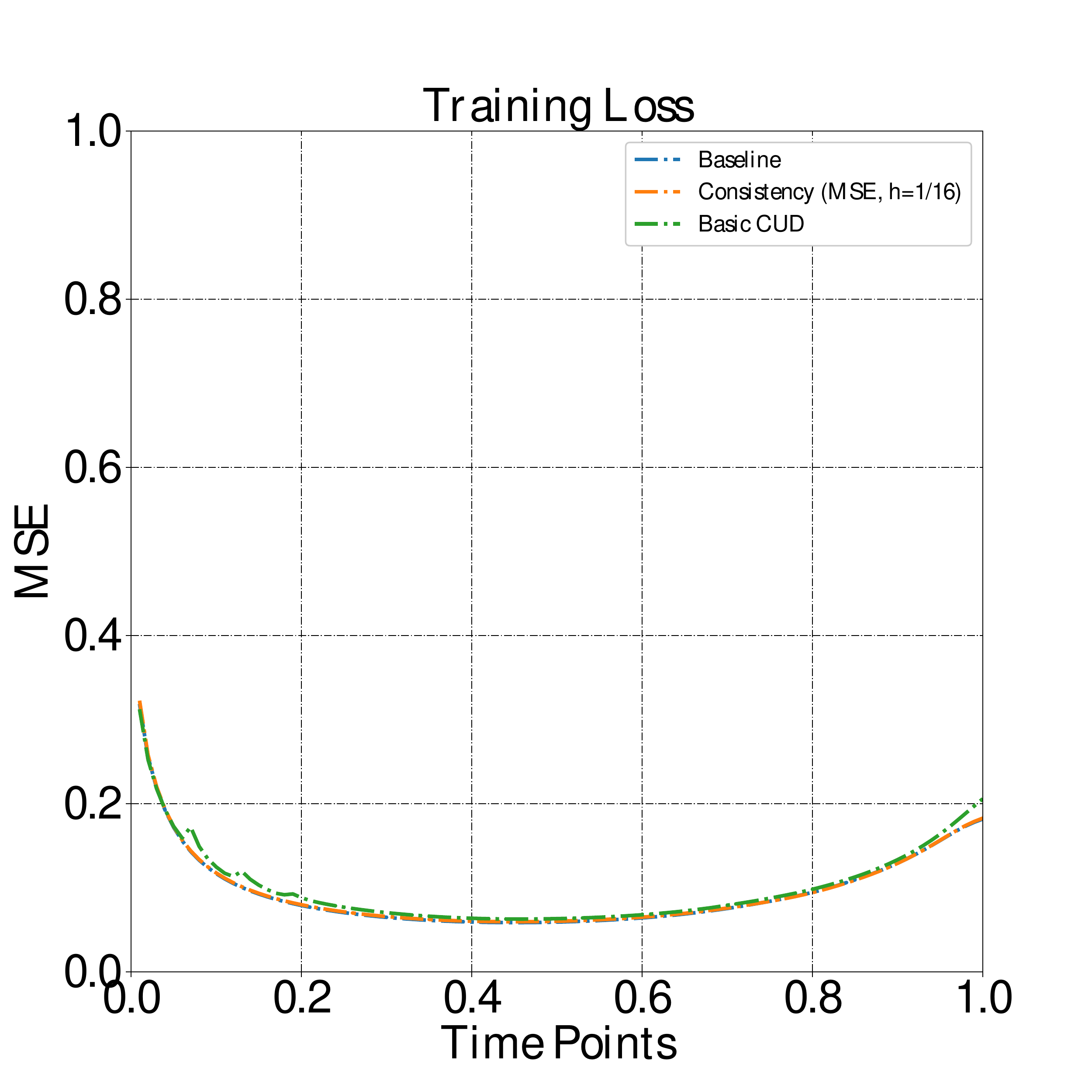}
\caption{Expectation of training losses at different time points.}
\label{fig:cost_fo_fit_analysis}
\vspace{-20pt}
\end{wrapfigure}
\section{Cost-of-Fit Analysis for Scenarios with Different $t$}
\label{apd:cost_fo_fit_analysis}
The optimization objective of DPMs is very different from the usual deep learning optimization objective, and one of the key differences is that DPMs has an additional input: the time point $t$. Ideally, the velocity estimation models corresponding to the different time points can form a set of functions: $\{f_\theta(\cdot,t)\}_{t}$. The number of neural network parameters is so large that most studies today apply weight sharing for velocity estimation models at different time steps, thus reducing storage costs. This approach also poses a problem at another level, in that at the end of the training phase, the expectation of training losses of the velocity estimation model at different time points varies considerably, as illustrated in Fig.~\ref{fig:cost_fo_fit_analysis}. Specifically, the expectation of the training loss will be greater as $t\rightarrow \epsilon$ or $t\rightarrow 1$, and smaller as $t\rightarrow 0.5$. This result is in line with our expectations, because as $t\rightarrow \epsilon$, the input is essentially free of Gaussian noise and can be approximated as $X_0$, but the output has to be predicted as $X_1-X_0$, when $X_1$ is completely unknowable and therefore extremely difficult. Similarly, as $t\rightarrow1$, the input is essentially Gaussian noise and can be approximated as $X_1$, but the output has to be predicted as $X_1-X_0$, when $X_0$ is completely unknowable and therefore also equally difficult.
\begin{figure*}[!h]
\includegraphics[width=1\textwidth,trim={0cm 3cm 0cm 5cm},clip]{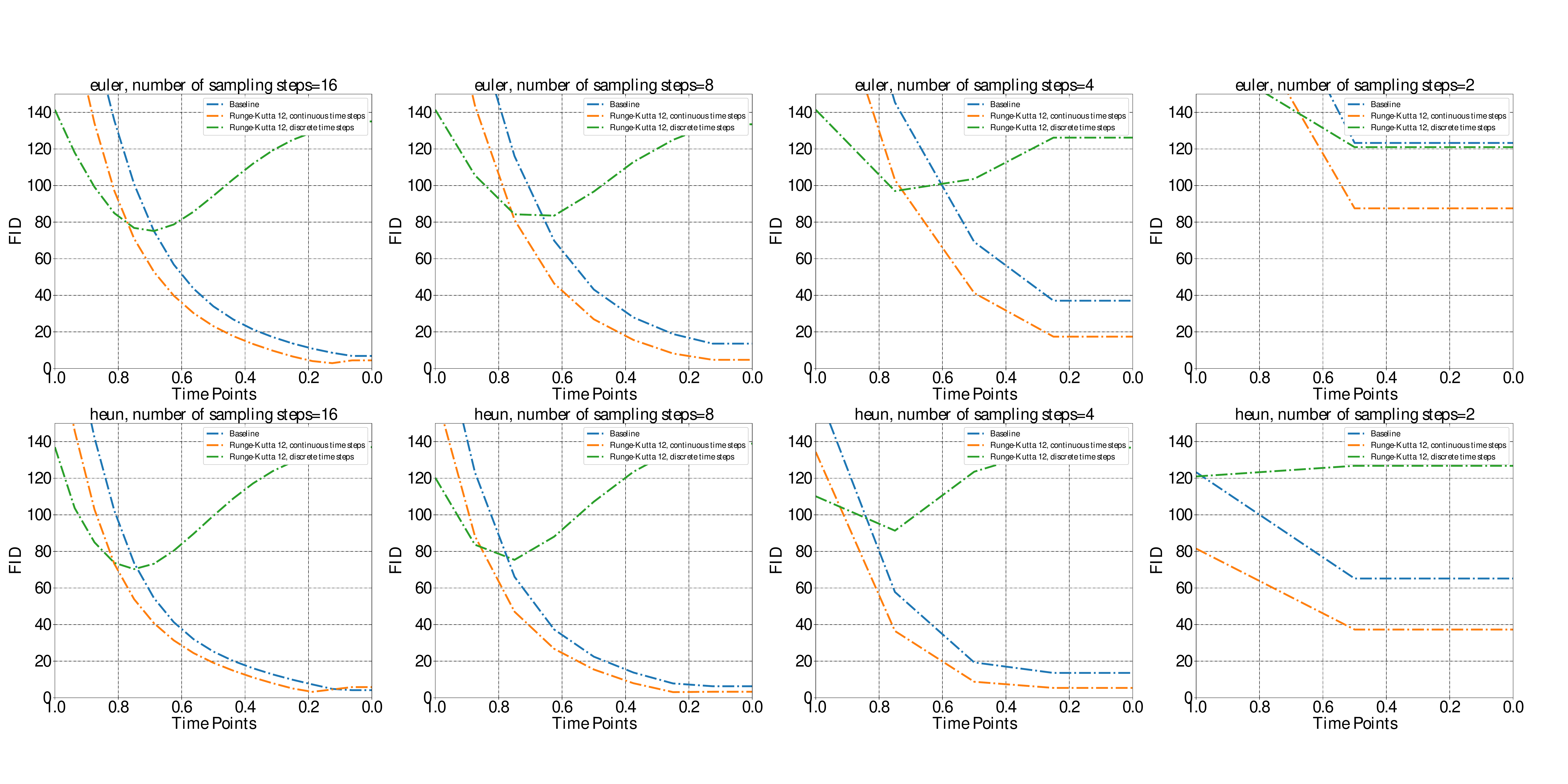}
\caption{The figure demonstrates that replacing continuous time steps with discrete time steps to perform CUD is not a feasible approach. The discrete time step is represented by $t$ ranging from $\frac{1}{N}$ to $\frac{N}{N}$, where $N$ is set to 16 by default. On the other hand, the continuous time step is represented by $t$ ranging from $\epsilon$ to 1.}
\label{fig:if_appling_discrete_training}
\vspace{-9pt}
\end{figure*}
\section{Final Multi-Step Distillation}
\label{sec:final_multi_step_distillation}
\renewcommand\arraystretch{0.95}
\setlength\tabcolsep{15pt}
\begin{table}[!h]
\footnotesize
\center
\caption{Ablation experiments on final multi-step distillation. \textbf{One-step}: vanilla one-step distillation technique introduced in~\cite{iclr22_rect}. \textbf{Multi-step}: Our proposed final multi-step distillation without SAM. \textbf{Multi-step+SAM}: Our proposed final multi-step distillation with SAM.}
\label{tab:final_multi_step_distillation_1}
\begin{tabular}{l|l|ccc}\toprule
\multicolumn{1}{c|}{\multirow{1}{*}{Dataset}} &  \multicolumn{1}{c|}{Metric} & \multicolumn{1}{c}{One-step} & \multicolumn{1}{c}{Multi-step}  & \multicolumn{1}{c}{Multi-step+SAM} \\\hline
\multirow{2}{*}{CIFAR-10} & FID($\downarrow$) & 5.78 & 3.74 & 3.37 \\
 & IS($\uparrow$) & 9.43 & 9.21 & 9.42 \\\hline
\multirow{2}{*}{MNIST} & FID($\downarrow$) & 8.94 & 4.80 & 6.38 \\
 & IS($\uparrow$) & 2.08 & 2.10 & 2.08 \\
\bottomrule
\end{tabular}
\end{table}
\renewcommand\arraystretch{0.95}
\setlength\tabcolsep{19pt}
\begin{table}[!h]
\footnotesize
\center
\caption{Verification of FID and IS at the end of each distillation step in final multi-step distillation. \textbf{FID}: FID is evaluated at the end of the current step. \textbf{IS}: IS is evaluated at the end of the current step. \textbf{Original FID}: The FID of the synthetic images at a specific step, is acquired through pre-training the model. And the synthetic images serve as the ground truth label, which needs to be aligned by final multi-step distillation.}
\label{tab:final_multi_step_distillation_2}
\begin{tabular}{l|l|ccc}\toprule
\multicolumn{1}{c|}{\multirow{1}{*}{Dataset}} &  \multicolumn{1}{c|}{Step} & \multicolumn{1}{c}{FID($\downarrow$)} & \multicolumn{1}{c}{Original FID($\downarrow$)}  & \multicolumn{1}{c}{IS($\uparrow$)} \\\hline
\multirow{3}{*}{CIFAR-10} & 8 & 24.84 & 23.12 & 8.61 \\
 & 11 & 14.36 & 9.72 & 9.25 \\
 & 14 & 3.37 & 2.91 & 9.42 \\
 \hline
\multirow{3}{*}{MNIST} & 8 & 12.59 & 13.71 & 2.03 \\
 & 11 & 7.56 & 4.67 & 2.04 \\
 & 14 & 6.38 & 2.80 & 2.08 \\
\bottomrule
\end{tabular}
\end{table}
Although CUD can accelerate sampling within a single training session, its reliance on continuous time steps necessitates fitting more time points than discrete time steps to complete the training, resulting in inferior performance compared to CD. In our experiments, merely replacing continuous time steps with discrete ones causes the model to collapse, as illustrated in Fig.~\ref{fig:if_appling_discrete_training}. Thus, we adopt the distillation technique (\textit{a.k.a.}, one-step distillation) from~\cite{iclr22_rect} to fit the velocity estimation model under a single time point, enabling the model to outperform CD in a one-step sampling scenario. Importantly, we incorporate ideas from the teacher-assistant concept to enhance the original distillation technique~\cite{mirzadeh2020improved}. Viewing the sample obtained in the last sampling step as a strong teacher output, the penultimate and penultimate third steps can be considered outputs from slightly weaker teachers. We can allow the model to gradually transition from an alignment time step from $1$ to $0$, effectively preventing performance degradation due to the gap between the strong teacher and the weak student.

For instance, if we have sampled 16 steps using Euler's method, we obtain a set of ``clean'' images for each sampling time point. We can first distill the model based on the ``clean'' images obtained in step 8, followed by those from steps 11 and 14. To ensure the model acquires a flat loss landscape after training, we align the training model's output with the EMA model's output, an approach interpretable as Sharpness-Aware Minimization (SAM)~\cite{nips2022_sam}. In final multi-step distillation, aligning the outputs of the training model and the EMA model does not always yield successful results. Thereby, we regard this approach as an optional method.

In our experiments, we employ final multi-step distillation using ``clean'' images obtained at steps 8, 11, and 14 with a sampling of 16 steps through Euler's method. To confirm the enhanced efficiency of our method compared to one-step distillation, we conduct comparative experiments and present the results in Table~\ref{tab:final_multi_step_distillation_1}. Final multi-step distillation demonstrates a significant improvement in FID evaluation compared to one-step distillation, indicating the effectiveness and reasonableness of the multi-stage guided distillation. The positive impact of SAM is less evident and only boosts IS without affecting FID. Furthermore, the effectiveness of each step in final multi-step distillation is illustrated in Table~\ref{tab:final_multi_step_distillation_2}. By guiding the training model through distillation using ``clean'' images from different steps, the synthetic images produced by the training model progressively improve in quality.

\section{Discussion}
\label{sec:limitation}
\paragraph{Limitation.} The CUD algorithm, which requires no pre-trained weights and a single training session, has considerably enhanced sampling acceleration compared to traditional DPM training paradigms. However, when assessed on the FID, a discernible performance gap with two-stage distillation-based accelerated sampling algorithms persists. This shortfall arises from two main factors: (1) Unlike those two-stage counterparts, which are trained on discrete time steps and necessitate a minimal number of time points to be fitted, CUD optimizes the loss function based on continuous time steps, demanding an infinite number of time points to be fitted. Yet, applying CUD directly to discrete time steps, as illustrated in Fig.~\ref{fig:if_appling_discrete_training}, leads to training collapse due to instability. (2) CUD's training duration is significantly shorter than that of two-stage distillation-based accelerated sampling algorithms.

Addressing this issue is a future work. This will involve revisiting hyperparameter selection and incorporating regularization terms to enhance training stability, thus enabling CUD to maintain stability over discrete time steps.

\paragraph{Broader Impact.} The idea of CUD is straightforward. It posits that the velocity estimation model output should align with the output from the same model at the previous moment, while simultaneously maintaining alignment with the ground truth label. This principle is anticipated to find broad application in future DPM training, potentially emerging as a more generalized paradigm. Consequently, we posit that CUD will exert a predominantly positive, rather than negative, impact on the community.

\section{Additional Experimental Results}
\label{apd:additional_experimental_results}
In this section, we present a series of experimental results that can not fit in the main paper due to space constraints, including Tables~\ref{tab:ablation_study_result_1_appendix}, ~\ref{tab:ablation_study_result_1_appendix_1},~\ref{tab:comparsion_results_appendix1},~\ref{tab:comparsion_results_appendix2},~\ref{tab:comparsion_results_about_solver}, and Fig.~\ref{fig:ablation_study_order_runge_kutta}. Tables~\ref{tab:ablation_study_result_1_appendix} and~\ref{tab:ablation_study_result_1_appendix_1} supplement the ablation experiments for Basic CUD in the main paper, while Tables~\ref{tab:comparsion_results_appendix1},~\ref{tab:comparsion_results_appendix2}, and~\ref{tab:comparsion_results_about_solver} supplement the comparison experimental results in the main paper. In particular, Tables~\ref{tab:comparsion_results_appendix1},~\ref{tab:comparsion_results_appendix2} are evaluated over samples sampled by the classical solver, including Euler and Heun, and Table~\ref{tab:comparsion_results_about_solver} is evaluated over samples sampled by \textit{training-free} accelerated sampling solvers, including DPM-solver-2 and Deis-2~\cite{zhang2023fast}.

We investigate the necessity of ground truth label supervision in this paragraph, showcasing both the Runge-Kutta-based multi-step alignment distillation that incorporates ground truth label supervision and its counterpart without such supervision in Fig.~\ref{fig:ablation_study_order_runge_kutta}. The label ``Runge-Kutta 23 (w/o GT)'' signifies CUD lacking the loss term $\textrm{MSE}(\widetilde{X}1-X_0,g_{\psi 2}(f_\theta(X_t,t)))$, while ``Runge-Kutta 23 (with GT)'' denotes the same algorithm but includes this loss term. Similarly, ``Runge-Kutta 34 (w/o GT)'' indicates CUD without the loss terms $\textrm{MSE}(\widetilde{X}_1-X_0,g{\psi 2}(f_\theta(X_t,t)))$ and $\textrm{MSE}(\widetilde{X}1-X_0,g{\psi 3}(f_\theta(X_t,t)))$, and ``Runge-Kutta 23 (with GT)'' implies the inclusion of these loss terms. The ground truth supervision is essential for CUD, as it helps circumvent improper gradient updates that could arise from unsupervised distillation and thus adversely affect performance. As a result, for $\mathcal{L}_\textrm{base}$ in the main paper, we incorporate ground truth supervision into each head of the velocity estimation model $f——\theta(\cdot,\cdot)$.

\ablationrungekutta
\renewcommand\arraystretch{1.1}
\setlength\tabcolsep{3pt}
\footnotesize
\begin{table}[h]
\center
\caption{Extension of Table~\ref{tab:ablation_study_result_1}. Additional experimental results on IS metric.}
\label{tab:ablation_study_result_1_appendix}
\resizebox{1.\textwidth}{!}{%
\begin{tabular}{ll|r|r|r|r|r|r|r|r}\toprule
\multicolumn{2}{c|}{\multirow{1}{*}{Methods}} &  \multicolumn{1}{c|}{RF (Eq.~\ref{eq:curvature_ode})} & \multicolumn{3}{c|}{Consistency Model (CD)} & \multicolumn{4}{c}{Basic CUD} \\\hline
\multicolumn{2}{c|}{\multirow{2}{*}{Settings}} &\multicolumn{1}{c|}{ \multirow{2}{*}{MSE}} & \multicolumn{1}{c|}{\multirow{2}{*}{MSE, $h\rightarrow 0$}} & \multicolumn{1}{c|}{\multirow{2}{*}{MSE, $h\!=\!1/16$}}  & \multicolumn{1}{c|}{\multirow{2}{*}{LPIPS,$h\!=\!1/16$}} &   \multicolumn{1}{c|}{MSE,$h\!=\!1/16$,}  & \multicolumn{1}{c|}{LPIPS,$h\!=\!1/16$,} &  \multicolumn{1}{c|}{MSE,$h\!=\!1/16$,} &  \multicolumn{1}{c}{LPIPS,$h\!=\!1/16$,} \\
& & \multicolumn{1}{c|}{}& \multicolumn{1}{c|}{} & \multicolumn{1}{c|}{} & \multicolumn{1}{c|}{} & \multicolumn{1}{c|}{vanilla weight} & \multicolumn{1}{c|}{vanilla weight} & \multicolumn{1}{c|}{dynamic weight} & \multicolumn{1}{c}{dynamic weight} \\\hline
\multicolumn{1}{c|}{\multirow{4}{*}{IS}} & Euler, 4 &  7.07 & 7.43 & 1.32 & 1.18 & 9.22 & 1.18 &  4.41 &  1.19\\
\multicolumn{1}{c|}{} & Euler, 16 & 8.85 & 9.19 &  1.33 &  1.19 & 10.20 & 1.18 & 3.98 & 1.18\\
\multicolumn{1}{c|}{} & Heun, 4 & 8.50 & 8.73 &  1.33 & 1.19 & 9.52 & 1.18 & 3.76 & 1.18 \\
\multicolumn{1}{c|}{} & Heun, 16 & 9.16 & 9.43 & 1.33 & 1.19 & 9.94 & 1.18 & 3.87 & 1.18 \\
\bottomrule
\end{tabular}}
\end{table}
\renewcommand\arraystretch{1.1}
\setlength\tabcolsep{3pt}
\footnotesize
\begin{table}[!h]
\center
\caption{Extension of Table~\ref{tab:ablation_study_result_1}. Additional experimental results on the loss function ``ML''. ``ML'' denotes ``MSE+LPIPS'', \textit{i.e.}, $\omega_1(i) \textrm{LPIPS}(\widetilde{X}_{t-h},f_\theta(X_t,t)) +\omega_2(i)\textrm{MSE}(\widetilde{X}_1-X_0,f_\theta(X_t,t))$.}
\label{tab:ablation_study_result_1_appendix_1}
\begin{tabular}{ll|r|r|r}\toprule
\multicolumn{2}{c|}{\multirow{1}{*}{Methods}} &  \multicolumn{1}{c|}{RF (Eq.~\ref{eq:curvature_ode})} & \multicolumn{2}{c}{Basic CUD} \\\bottomrule
\multicolumn{2}{c|}{\multirow{2}{*}{Settings}} &  \multicolumn{1}{c|}{\multirow{2}{*}{MSE}} & \multicolumn{1}{c|}{ML,$h\!=\!1/16$,}  & \multicolumn{1}{c}{ML,$h\!=\!1/16$,}\\
 \multicolumn{2}{c|}{} &  \multicolumn{1}{c|}{} & \multicolumn{1}{c|}{vanilla weight} & \multicolumn{1}{c}{dynamic weight}\\ \hline
 \multicolumn{1}{c|}{\multirow{4}{*}{FID}} & Euler, 4 & 37.05& \textcolor{C1}{171.56} & \textcolor{C1}{302.70}  \\
\multicolumn{1}{c|}{} & Euler, 16& 6.85& \textcolor{C1}{129.83} & \textcolor{C1}{275.02} \\
\multicolumn{1}{c|}{} & Heun, 4& 13.58& \textcolor{C1}{130.10} & \textcolor{C1}{270.85} \\
\multicolumn{1}{c|}{}  & Heun, 16& 4.15& \textcolor{C1}{128.48} & \textcolor{C1}{340.82} \\\hline
\multicolumn{1}{c|}{\multirow{4}{*}{IS}} & Euler, 4& 7.07 & 3.45 & 1.75  \\
\multicolumn{1}{c|}{} & Euler, 16& 8.85 & 3.54 & 1.86 \\
\multicolumn{1}{c|}{} & Heun, 4& 8.50 & 3.54 &  1.91 \\
\multicolumn{1}{c|}{}  & Heun, 16& 9.16 & 3.55 & 1.69 \\
\bottomrule
\end{tabular}
\end{table}
\begin{table*}
    \begin{minipage}[t]{0.49\linewidth}
	\caption{Additional experimental results leveraging Euler and Heun on CIFAR-10.}\label{tab:comparsion_results_appendix1}
	\centering
	{\setlength{\extrarowheight}{1.5pt}
	\begin{adjustbox}{max width=\linewidth}
	\begin{tabular}{lcccc}
        \Xhline{3\arrayrulewidth}
        \Xhline{3\arrayrulewidth}
	    METHOD & Solver & NFE ($\downarrow$) & FID ($\downarrow$) & IS ($\uparrow$) \\
      \\[-2ex]
        \multicolumn{4}{l}{\textbf{Diffusion Model (One Stage)}}\\\Xhline{3\arrayrulewidth}
        Curvature~\cite{icml23_curvature} & Euler & 16 &  6.85 & 8.84 \\
        Curvature & Euler & 4 &  37.05 & 7.07 \\
        Curvature & Heun & 7 &  13.58 & 8.49 \\   
        Curvature & Heun & 31 &  4.15 & 9.16 \\
        CUD (Runge-Kutta 12) & Euler & 15 &  2.91 & 9.90 \\
        CUD (Runge-Kutta 12) & Euler & 4 &  17.40 & 9.38 \\
        CUD (Runge-Kutta 12) & Heun & 7 &  5.38 & 9.61 \\
        CUD (Runge-Kutta 12) & Heun & 31 &  3.24 & 9.75 \\
        CUD (Runge-Kutta 23) & Euler & 15 & 2.80 & 9.36 \\
        CUD (Runge-Kutta 23) & Euler & 4 & 12.23 & 8.39 \\
        CUD (Runge-Kutta 23) & Heun & 7 & 5.07 & 9.20 \\
        CUD (Runge-Kutta 23) & Heun & 29 & 4.49 & 9.45 \\
        CUD (Runge-Kutta 34) & Euler & 14 & 3.40 & 9.39 \\
        CUD (Runge-Kutta 34) & Euler & 4 & 9.45 & 8.50 \\
        CUD (Runge-Kutta 34) & Heun & 7 & 4.70 & 9.34 \\
        CUD (Runge-Kutta 34) & Heun & 25 & 4.38 & 9.66 \\
        2-Rectified flow$^*$ & Euler & 1 & 4.85 & 9.01 \\\Xhline{3\arrayrulewidth}
	\end{tabular}
    \end{adjustbox}
	}
\end{minipage}
\hfill
\begin{minipage}[t]{0.49\linewidth}
	\caption{Additional experimental results leveraging Euler and Heun on MNIST.}\label{tab:comparsion_results_appendix2}
	\centering
	{\setlength{\extrarowheight}{1.5pt}
	\begin{adjustbox}{max width=\linewidth}
	\begin{tabular}{lcccc}
	    \Xhline{3\arrayrulewidth}
        \Xhline{3\arrayrulewidth}
	    METHOD & Solver & NFE ($\downarrow$) & FID ($\downarrow$) & IS ($\uparrow$) \\
      \\[-2ex]
        \multicolumn{5}{l}{\textbf{MNIST (One Stage)}}\\\Xhline{3\arrayrulewidth}
        Curvature~\cite{icml23_curvature} & Euler & 16 & 21.77 & 2.11 \\
        Curvature & Euler & 4 & 49.96 & 2.02 \\
        Curvature & Heun & 7 & 21.55 & 2.10 \\
        Curvature & Heun & 31 & 18.56 & 2.12 \\
        CUD (Runge-Kutta 12) & Euler & 16 & 3.39 & 2.01 \\
        CUD (Runge-Kutta 12) & Euler & 4 & 9.45 & 2.02 \\
        CUD (Runge-Kutta 12) & Heun & 31 & 3.49 & 2.12 \\
        CUD (Runge-Kutta 12) & Heun & 7 & 3.06 & 2.05 \\
        CUD (Runge-Kutta 23) & Euler & 16 & 2.18 & 2.10 \\
        CUD (Runge-Kutta 23) & Euler & 4 & 7.06 & 2.03 \\
        CUD (Runge-Kutta 23) & Heun & 31 & 2.37 & 2.12 \\
        CUD (Runge-Kutta 23) & Heun & 7 & 2.81 & 2.05 \\
        CUD (Runge-Kutta 34) & Euler & 16 & 1.81 & 2.10 \\
        CUD (Runge-Kutta 34) & Euler & 4 & 6.70 & 2.03 \\
        CUD (Runge-Kutta 34) & Heun & 31 & 1.87 & 2.12 \\
        CUD (Runge-Kutta 34) & Heun & 7 & 2.87 & 2.05 \\
        PD  & Euler & 1 & 13.64 & 2.18 \\\Xhline{3\arrayrulewidth}
	\end{tabular}
    \end{adjustbox}
    }
\end{minipage}
\end{table*}
 \renewcommand\arraystretch{1.1}
\setlength\tabcolsep{3pt}
\footnotesize
\begin{table*}[t]
	\caption{Additional experimental results leveraging DPM-solver-2 and Deis-2 on CIFAR-10.}\label{tab:comparsion_results_about_solver}
	\centering
 \setlength{\extrarowheight}{1.5pt}
 \resizebox{1.\textwidth}{!}{%
	\begin{tabular}{c|cc|cc|cc|cc}
        \Xhline{3\arrayrulewidth}
	    \multirow{3}{*}{METHOD} & \multicolumn{8}{c}{Solver} \\\cline{2-9}
       & \multicolumn{2}{c|}{DPM-solver-2 (NFE=7)} & \multicolumn{2}{c|}{DPM-solver-2 (NFE=4)} & \multicolumn{2}{c|}{Deis-2 (NFE=7)} &  \multicolumn{2}{c}{Deis-2 (NFE=4)}\\
     & FID ($\downarrow$) & IS ($\uparrow$) & FID ($\downarrow$) &  IS ($\uparrow$) &  FID ($\downarrow$) &  IS ($\uparrow$) & FID ($\downarrow$) &  IS ($\uparrow$) \\\hline
CUD (Runge-Kutta 12) & 3.27 & 9.54 &  8.25 & 8.95 & 4.70 &  9.76 & 8.53 & 9.15 \\
CUD (Runge-Kutta 23) & 3.53 & 9.35 &  6.42 & 8.90 & 4.90 & 9.43 & 6.30 & 8.99 \\
CUD (Runge-Kutta 34) & 4.55 & 9.33 &  6.97 & 8.95 & 6.30& 8.99 & 5.35 & 9.35 \\
       \Xhline{3\arrayrulewidth}
	\end{tabular}}
\end{table*}
\section{Implementation Details}
\label{apd:implementation_details}
Table.~\ref{tab:config_all} shows the training and architecture configuration we use in our experiments. For CIFAR-10, MNIST, and ImageNet 64$\times$64 datasets, we carry out three different configurations (\textit{i.e.}, (a), (b), and (c)) for experiments, where configurations (a) and (b) are following~\cite{icml23_curvature} and the model architecture in configuration (c) is following~\cite{icml23_consistency}. We ran the code on NVIDIA Tesla A100 GPUs, where configurations (a) used 4 GPUs with a batch size of 32 on each GPU. Configuration (b) applied 4 GPUs with a batch size of 64 on each GPU. Configuration (c) applied 8 GPUs with a batch size of 128 on each GPU.  Configuration (d) is used for the final multi-step distillation, and its batch size and the learning rate are strongly correlated with the dataset. For example, if the final discrete distillation is performed on MNIST, then its batch size and learning rate are the same as configuration (b).
\begin{table}[!h]
\centering
\caption{Architecture and training configurations on CIFAR-10, MNIST and ImageNet 64$\times$64.}
\vskip 0.15in
\begin{tabular}{@{}lcccc@{}}
\toprule
                         & CIFAR-10 (a)         & MNIST (b)         & ImageNet 64$\times$64 (c)          & Final Multi-Step Distillation (d)      \\ \midrule
Iterations               &  500k                &   50k               & 500k                & 40k$\times$3               \\
Batch size               & 128                  & 256                  & 1024                & 128/256/-              \\
Learning rate            & 2e-4               & 3e-4               & 2e-4         & 2e-4/3e-4/-        \\
LR warm-up steps         & 5000                 & 8000                 & 0               & 0              \\
EMA decay rate           & 0.9999               & 0.9999               & 0.9999             & 0.9999            \\
EMA start steps          & 1                    & 300                    & 1               & 1             \\
Dropout probability      & 0                 & 0.13                & 0.0              & 0.0            \\
Channel multiplier       & 128                  & 32                  & 192                 &  -             \\
Channels per resolution  & $[2, 2, 2]$          & $[2, 2, 2]$          & $[1,2,3,4]$        &  -    \\
Xflip augmentation       & 0                    & X                    & X                  & X                 \\
\# of params (generator) & 55.73M               & 2.2M               & 295.90M             & -           \\
\# of params (encoder)   & 2.2M                 & 2.2M                 & -              & -              \\
\# of ResBlocks          & 4                    & 2                    & 3                 & -         \\
$t$ range                & $[1e-5,1]$           & $[1e-5, 1]$       & $[1e-5, 1]$        & -          \\
                         \bottomrule
\end{tabular}
\label{tab:config_all}
\end{table}

\section{Additional Content Presentation}
\label{apd:additional_data_presentation}
When $t\rightarrow \epsilon$, the comparison between the curves in Figs.~\ref{fig:ablation_study_if_use_ema} and~\ref{fig:ablation_study_if_use_random_step_and_shakedrop} in the main paper are not fuzzy. Therefore, we present the specific data in Table~\ref{tab:table_ema_cud} and~\ref{tab:table_random_step_size}. These tables are labeled in exactly the same form and with the same data content as Figs.~\ref{fig:ablation_study_if_use_ema} and~\ref{fig:ablation_study_if_use_random_step_and_shakedrop}.

\renewcommand\arraystretch{1.1}
\setlength\tabcolsep{3pt}
\footnotesize
\begin{table}[h]
\center
\caption{Ablation experiments on whether to use the EMA model for CUD.}
\label{tab:table_ema_cud}
\resizebox{1.\textwidth}{!}{%
\begin{tabular}{l|l|l|l|rrrrrrrrrrrrrrrr}\toprule
\multirow{2}{*}{Method} & \multirow{2}{*}{$\hat{h}$} & \multirow{2}{*}{N}& \multirow{2}{*}{ODE Solver} &  \multicolumn{16}{c}{Time Points} \\\cline{5-20}
&&&& $\frac{16}{16}$ & $\frac{15}{16}$ & $\frac{14}{16}$ & $\frac{13}{16}$ & $\frac{12}{16}$ & $\frac{11}{16}$ & $\frac{10}{16}$ & $\frac{9}{16}$ & $\frac{8}{16}$ & $\frac{7}{16}$ & $\frac{6}{16}$ & $\frac{5}{16}$ & $\frac{4}{16}$ & $\frac{3}{16}$ & $\frac{2}{16}$ & $\frac{1}{16}$ \\\hline
\multirow{8}{*}{Baseline} &  \multirow{8}{*}{$\frac{1}{16}$} & 16 & Euler&292.58&225.57&178.35&135.57&100.76&75.11&56.83&43.61&33.97&26.83&21.34&17.15&13.68&10.86&8.59&6.85 \\
&  & 8 & Euler &  292.58&&188.48&&116.05&&69.85&&43.18&&27.91&&18.83&&13.58 & \\
&  & 4 & Euler &  292.58&&&&145.44&&&&69.17&&&&37.05&&& \\
&  & 2 & Euler &292.58&&&&&&&&123.29&&&&&&& \\
&  & 16 & Heun &245.83&191.65&142.16&102.29&73.65&54.48&41.33&32.01&25.24&20.15&16.12&12.74&9.84&7.28&4.84&4.16 \\
& & 8 & Heun &214.08&&123.26&&66.11&&37.31&&22.47&&13.79&&7.83&&6.29& \\
& & 4 & Heun &170.07&&&&57.80&&&&19.32&&&&13.58&&& \\
& & 2 & Heun &123.29&&&&&&&&65.18&&&&&&&\\\bottomrule
\multirow{8}{*}{\rowtl{Runge-Kutta 12}{($f_{\theta_-}$)}} &  \multirow{8}{*}{$\frac{1}{16}$} & 16 & Euler&138.92&101.45&73.48&55.80&44.54&36.79&31.10&26.49&22.82&18.72&15.39&12.67&10.58&7.88&8.94&13.46 \\
&  & 8 & Euler &  138.92&&75.92&&45.84&&30.39&&20.81&&13.54&&8.84&&7.56&\\
&  & 4 & Euler &  138.92&&&&51.72&&&&21.08&&&&7.92&&& \\
&  & 2 & Euler &138.92&&&&&&&&34.27&&&&&&&\\
&  & 16 & Heun &114.15&80.80&60.33&48.32&40.59&34.97&30.44&26.43&22.69&19.21&16.20&13.73&10.60&10.19&14.85&17.65 \\
& & 8 & Heun &91.24&&51.83&&35.36&&25.81&&18.31&&12.66&&9.15&&10.76&\\
& & 4 & Heun &63.26&&&&23.82&&&&9.13&&&&7.57&&&\\
& & 2 & Heun &34.28&&&&&&&&15.75&&&&&&&\\\bottomrule
\multirow{8}{*}{\rowtl{Runge-Kutta 12}{($f_{\theta}$)}} &  \multirow{8}{*}{$\frac{1}{16}$} & 16 & Euler&170.70&128.28&93.18&67.27&49.31&37.09&28.58&22.20&16.89&12.34&8.58&5.84&4.25&4.09&7.19&11.08 \\
&  & 8 & Euler & 170.70&&99.65&&54.42&&30.70&&17.21&&8.10&&3.65&&6.96& \\
&  & 4 & Euler &63.26&&&&23.82&&&&9.13&&&&7.57&&&\\
&  & 2 & Euler & 170.70&&&&&&&&45.21&&&&&&&\\
&  & 16 & Heun &143.21&101.73&72.04&52.49&39.80&31.40&25.27&20.23&15.81&11.97&8.93&6.89&5.68&7.78&12.65&14.79 \\
& & 8 & Heun &117.71&&60.90&&33.92&&20.45&&11.52&&6.10&&6.77&&10.16& \\
& & 4 & Heun &81.76&&&&22.56&&&&4.30&&&&4.19&&&\\
& & 2 & Heun &45.21&&&&&&&&17.70&&&&&&&\\\bottomrule
\multirow{8}{*}{\rowtl{Runge-Kutta 23}{($f_{\theta_-}$)}} &  \multirow{8}{*}{$\frac{1}{16}$} & 16 & Euler &102.27&77.05&59.48&46.99&38.70&33.07&29.02&25.91&23.37&21.24&19.55&18.41&17.67&16.88&16.18&15.96 \\
&  & 8 & Euler &102.27&&61.37&&38.46&&26.84&&20.44&&16.46&&14.60&&13.11&\\
&  & 4 & Euler &102.27&&&&41.06&&&&17.90&&&&11.31&&& \\
&  & 2 & Euler &102.27&&&&&&&&23.25&&&&&&&\\
&  & 16 & Heun &85.37&64.52&51.04&42.63&37.21&33.37&30.39&27.91&25.73&23.92&22.63&21.84&21.33&20.77&20.45&20.38 \\
& & 8 & Heun &70.85&&44.12&&32.48&&26.29&&22.23&&20.10&&19.07&&18.99&\\
& & 4 & Heun &49.40&&&&21.68&&&&15.12&&&&15.75&&&\\
& & 2 & Heun &23.25&&&&&&&&13.14&&&&&&&\\\bottomrule
\multirow{8}{*}{\rowtl{Runge-Kutta 23}{($f_{\theta}$)}} &  \multirow{8}{*}{$\frac{1}{16}$} & 16 & Euler & 139.64&103.35&74.87&54.49&41.09&32.14&26.10&21.72&18.37&15.77&13.69&12.15&11.35&10.74&10.11&9.71\\
&  & 8 & Euler &139.64&&79.21&&41.22&&25.47&&16.29&&10.98&&8.35&&7.07&\\
&  & 4 & Euler &139.64&&&&50.69&&&&16.67&&&&6.70&&& \\
&  & 2 & Euler &139.63&&&&&&&&20.01&&&&&&&\\
&  & 16 & Heun &115.70&82.15&59.34&44.95&36.05&30.21&26.13&23.04&20.58&18.59&17.05&16.16&15.64&15.10&14.61&14.50 \\
& & 8 & Heun &93.76&&49.67&&30.53&&21.84&&16.90&&14.39&&13.38&&13.35&\\
& & 4 & Heun &62.75&&&&19.49&&&&9.75&&&&10.32&&&\\
& & 2 & Heun &29.01&&&&&&&&12.88&&&&&&&\\\bottomrule
\end{tabular}}
\end{table}

\clearpage
\renewcommand\arraystretch{1.1}
\setlength\tabcolsep{3pt}
\footnotesize
\begin{table}[h]
\center
\caption{Ablation experiments on whether to use the no-fix-step size and dynamic skip connection for CUD.}
\label{tab:table_random_step_size}
\resizebox{1.\textwidth}{!}{%
\begin{tabular}{l|l|l|l|rrrrrrrrrrrrrrrr}\toprule
\multirow{2}{*}{Method} & \multirow{2}{*}{$\hat{h}$} & \multirow{2}{*}{N}& \multirow{2}{*}{ODE Solver} &  \multicolumn{16}{c}{Time Points} \\\cline{5-20}
&&&& $\frac{16}{16}$ & $\frac{15}{16}$ & $\frac{14}{16}$ & $\frac{13}{16}$ & $\frac{12}{16}$ & $\frac{11}{16}$ & $\frac{10}{16}$ & $\frac{9}{16}$ & $\frac{8}{16}$ & $\frac{7}{16}$ & $\frac{6}{16}$ & $\frac{5}{16}$ & $\frac{4}{16}$ & $\frac{3}{16}$ & $\frac{2}{16}$ & $\frac{1}{16}$ \\\hline
\multirow{8}{*}{\rowtl{Runge-Kutta 12}{, uniform}} &  \multirow{8}{*}{$\frac{1}{16}$} & 16 & Euler & 235.29&179.87&133.11&96.53&70.32&52.03&38.97&29.51&23.49&17.13&12.87&9.32&6.31&3.95&3.46&4.79 \\
&  & 8 & Euler &235.29&&142.47&&80.62&&45.88&&26.44&&15.45&&7.99&&5.05&\\
&  & 4 & Euler &235.29&&&&102.27&&&&40.94&&&&17.64&&& \\
&  & 2 & Euler &235.29&&&&&&&&81.18&&&&&&&\\
&  & 16 & Heun &199.64&145.16&102.18&72.36&52.90&39.72&30.51&23.73&18.54&14.39&10.84&7.72&5.00&3.36&4.89&6.29 \\
& & 8 & Heun &167.76&&87.53&&46.16&&26.00&&14.97&&7.70&&3.17&&3.60&\\
& & 4 & Heun &124.11&&&&35.99&&&&8.56&&&&5.40&&&\\
& & 2 & Heun &81.18&&&&&&&&36.30&&&&&&&\\\bottomrule
\multirow{8}{*}{\rowtl{Runge-Kutta 12}{, rule}} &  \multirow{8}{*}{$\frac{1}{16}$} & 16 & Euler &361.05&278.28&173.90&100.28&63.26&46.67&37.97&30.96&23.84&17.60&13.30&10.66&8.92&7.52&6.64&9.19 \\
&  & 8 & Euler &361.05&&186.91&&71.69&&43.19&&27.07&&14.99&&10.54&&8.75&\\
&  & 4 & Euler &361.05&&&&90.06&&&&38.92&&&&18.95&&& \\
&  & 2 & Euler &361.05&&&&&&&&74.18&&&&&&&\\
&  & 16 & Heun &319.92&209.83&117.32&68.82&47.83&38.14&31.80&25.69&19.73&15.07&11.96&9.86&8.18&6.79&8.11&9.16 \\
& & 8 & Heun &260.24&&88.28&&42.62&&28.05&&17.58&&9.77&&6.89&&6.34&\\
& & 4 & Heun &146.51&&&&36.49&&&&10.63&&&&8.45&&&\\
& & 2 & Heun &74.18&&&&&&&&33.25&&&&&&&\\\bottomrule
\multirow{8}{*}{\rowtl{Runge-Kutta 23}{, uniform}} &  \multirow{8}{*}{$\frac{1}{16}$} & 16 & Euler &228.30&166.15&117.39&82.49&58.95&43.19&32.57&26.16&19.80&15.71&12.38&9.58&7.19&5.25&3.93&3.27 \\
&  & 8 & Euler &228.30&&126.41&&67.54&&37.53&&21.98&&13.04&&7.34&&4.24&\\
&  & 4 & Euler &228.30&&&&87.30&&&&32.45&&&&13.13&&&\\
&  & 2 & Euler &228.30&&&&&&&&67.08&&&&&&&\\
&  & 16 & Heun &187.60&129.34&87.81&61.15&44.52&33.93&26.83&21.83&18.02&14.90&12.61&9.70&7.55&5.89&4.93&4.80\\
& & 8 & Heun &151.39&&73.93&&38.51&&22.90&&14.67&&9.13&&5.29&&4.74&\\
& & 4 & Heun &107.19&&&&29.19&&&&8.24&&&&5.99&&&\\
& & 2 & Heun &67.08&&&&&&&&29.03&&&&&&&\\\bottomrule
\multirow{8}{*}{\rowtl{Runge-Kutta 23}{, rule}} &  \multirow{8}{*}{$\frac{1}{16}$} & 16 & Euler &233.71&159.26&108.23&74.60&53.46&39.97&30.86&24.31&19.54&16.02&13.32&11.05&8.92&6.74&4.74&4.11 \\
&  & 8 & Euler &233.71&&115.43&&59.76&&34.30&&20.84&&13.49&&8.64&&4.74&\\
&  & 4 & Euler &233.71&&&&74.01&&&&28.54&&&&12.61&&&\\
&  & 2 & Euler &233.71&&&&&&&&56.06&&&&&&&\\
&  & 16 & Heun &186.61&122.01&80.84&56.27&41.71&32.66&26.46&21.93&18.53&15.90&13.71&11.65&9.44&7.03&5.73&5.90\\
& & 8 & Heun &146.03&&66.79&&36.25&&22.78&&15.63&&11.00&&6.40&&5.56&\\
& & 4 & Heun &94.17&&&&24.04&&&&9.43&&&&7.55&&&\\
& & 2 & Heun &56.06&&&&&&&&24.83&&&&&&&\\\bottomrule
\multirow{8}{*}{\rowtl{Runge-Kutta 12}{, uniform, $c_\textrm{skip}=0.25$}} &  \multirow{8}{*}{$\frac{1}{16}$} & 16 & Euler&235.34&181.63&135.87&90.32&72.89&54.22&41.00&31.29&24.26&18.74&14.34&10.64&7.37&4.62&3.19&4.44 \\
&  & 8 & Euler 235.34&&145.96&&83.65&&48.05&&28.43&&16.80&&9.16&&5.08&\\
&  & 4 & Euler &235.34&&&&105.99&&&&43.40&&&&18.92&&&\\
&  & 2 & Euler &235.34&&&&&&&&84.16&&&&&&&\\
&  & 16 & Heun &200.50&147.38&104.67&74.76&54.88&41.53&32.18&25.31&20.03&15.76&12.11&8.84&5.86&3.73&4.77&6.17\\
& & 8 & Heun &170.74&&99.47&&48.28&&27.73&&16.42&&8.79&&3.44&&3.69&\\
& & 4 & Heun &128.56&&&&38.07&&&&9.62&&&&6.22&&&\\
& & 2 & Heun &84.16&&&&&&&&38.55&&&&&&&\\\bottomrule
\multirow{8}{*}{\rowtl{Runge-Kutta 12}{, uniform, $c_\textrm{skip}=0.75$}} &  \multirow{8}{*}{$\frac{1}{16}$} & 16 & Euler&237.09&180.73&134.04&97.13&70.98&52.86&39.80&30.26&23.12&17.70&13.39&9.72&6.56&4.04&\textbf{2.91}&4.33\\
&  & 8 & Euler & 237.09&&143.74[&&81.21&&46.42&&26.91&&15.60&&8.18&&4.74&\\
&  & 4 & Euler & 237.09&&&&103.04&&&&41.26&&&&17.40&&&\\
&  & 2 & Euler &237.09&&&&&&&&87.53&&&&&&&\\
&  & 16 & Heun &199.64&145.87&102.9&73.21&53.87&40.76&31.38&24.42&19.11&14.83&11.09&7.91&5.06&3.24&4.54&5.83\\
& & 8 & Heun &167.64&&88.37&&47.04&&26.75&&15.46&&7.94&&3.08&&3.29&\\
& & 4 & Heun &134.37&&&&36.49&&&&8.75&&&&5.38&&&\\
& & 2 & Heun &81.53&&&&&&&&37.29&&&&&&&\\\bottomrule
\end{tabular}}
\end{table}
\section{Additional Samples from Catch-Up Distillation and Final Multi-Step Distillation}
\label{apd:synthetic_image}
We provide additional samples from Catch-Up Distillation (CUD) and Final Multi-Step Distillation (FMSD) on MNIST (Figs.~\ref{fig:mnist_full}), CIFAR-10 (Figs.~\ref{fig:cifar10_full}), and ImageNet 64$\times$64 (Figs.~\ref{fig:imagenet_64_full}).
\begin{figure}[!b]
    \centering
    \begin{subfigure}[!b]{\textwidth}
        \includegraphics[width=\textwidth]{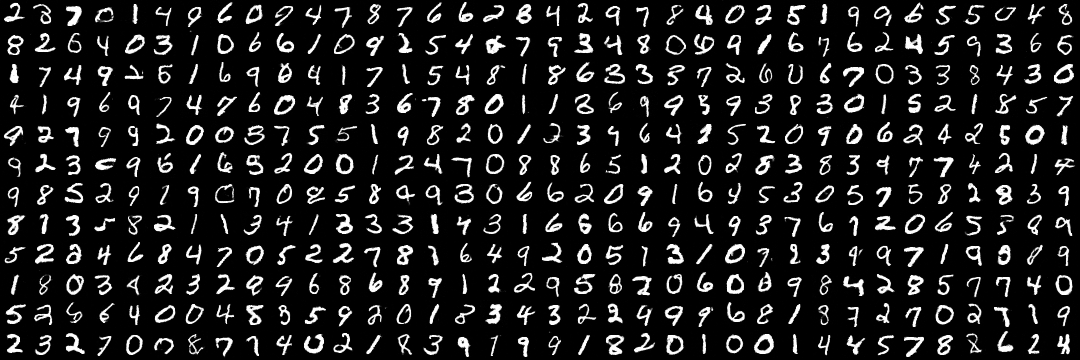}
        \caption{FMSD with 1 step (FID=6.36)}
    \end{subfigure}
    \begin{subfigure}[!b]{\textwidth}
        \includegraphics[width=\textwidth]{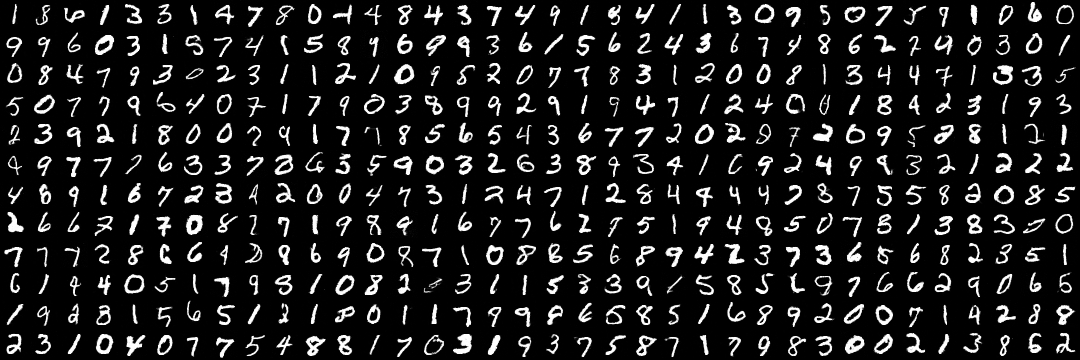}
        \caption{CUD Runge-Kutta 12 with 16 steps (FID=3.99)}
    \end{subfigure}
    \begin{subfigure}[!b]{\textwidth}
        \includegraphics[width=\textwidth]{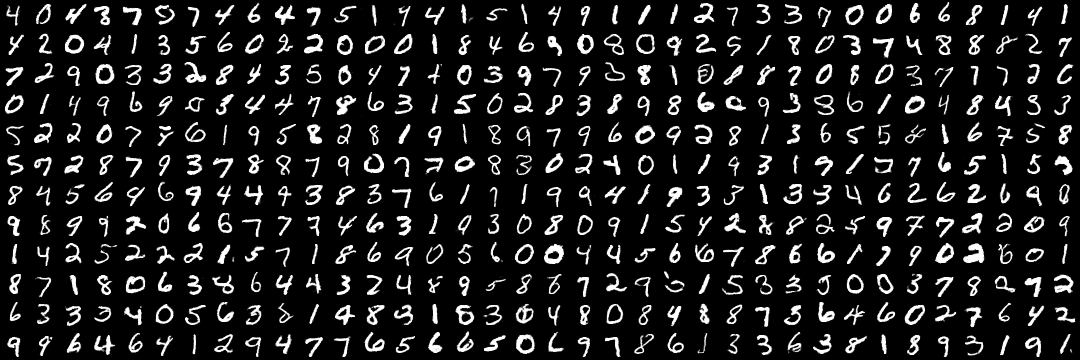}
        \caption{CUD Runge-Kutta 23 with 16 steps (FID=2.08)}
    \end{subfigure}
    \begin{subfigure}[!b]{\textwidth}
        \includegraphics[width=\textwidth]{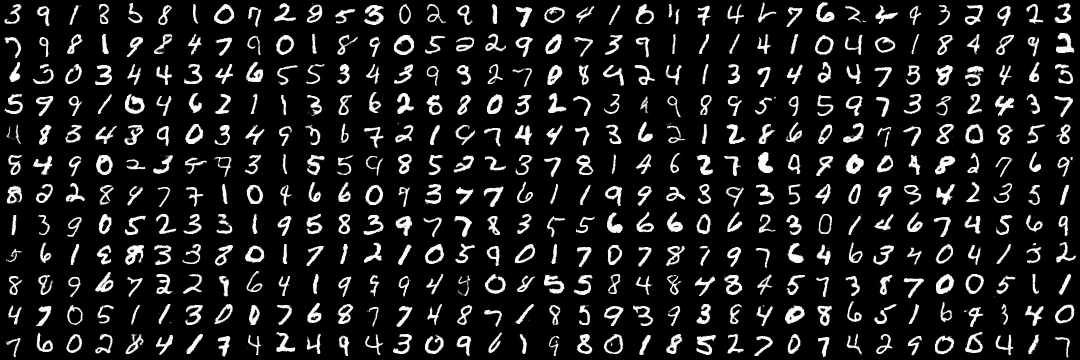}
        \caption{CUD Runge-Kutta 34 with 16 steps (FID=1.81)}
    \end{subfigure}
    \caption{Synthetic images sampled from MNIST $28\times 28$.}
    \label{fig:mnist_full}
\end{figure}
\begin{figure}[!b]
    \centering
    \begin{subfigure}[!b]{\textwidth}
        \includegraphics[width=\textwidth]{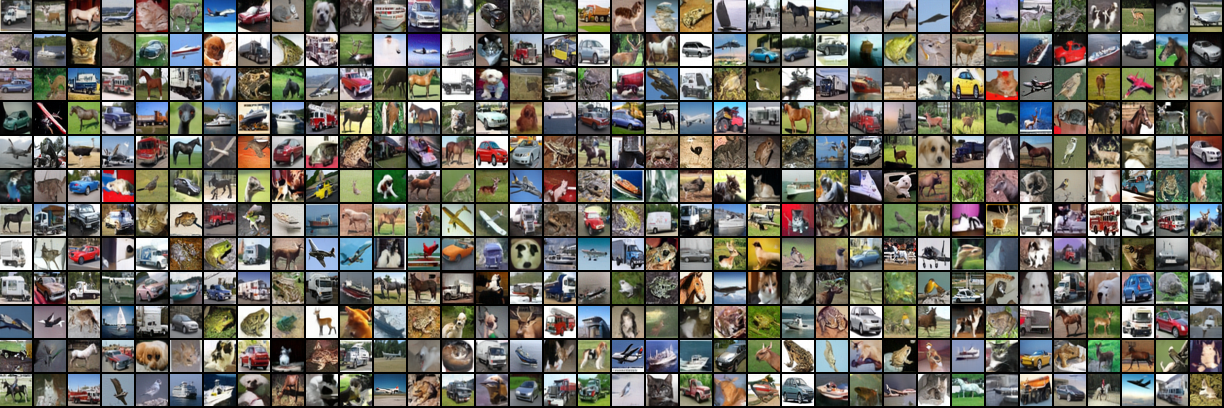}
        \caption{FMSD with 1 step (FID=3.77)}
    \end{subfigure}
    \begin{subfigure}[!b]{\textwidth}
        \includegraphics[width=\textwidth]{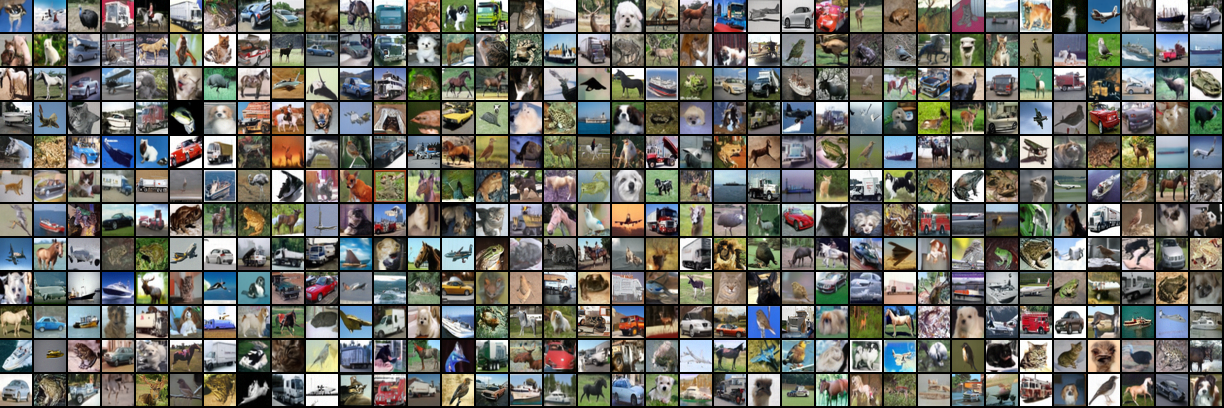}
        \caption{CUD Runge-Kutta 12 with 15 steps (FID=2.91)}
    \end{subfigure}
    \begin{subfigure}[!b]{\textwidth}
        \includegraphics[width=\textwidth]{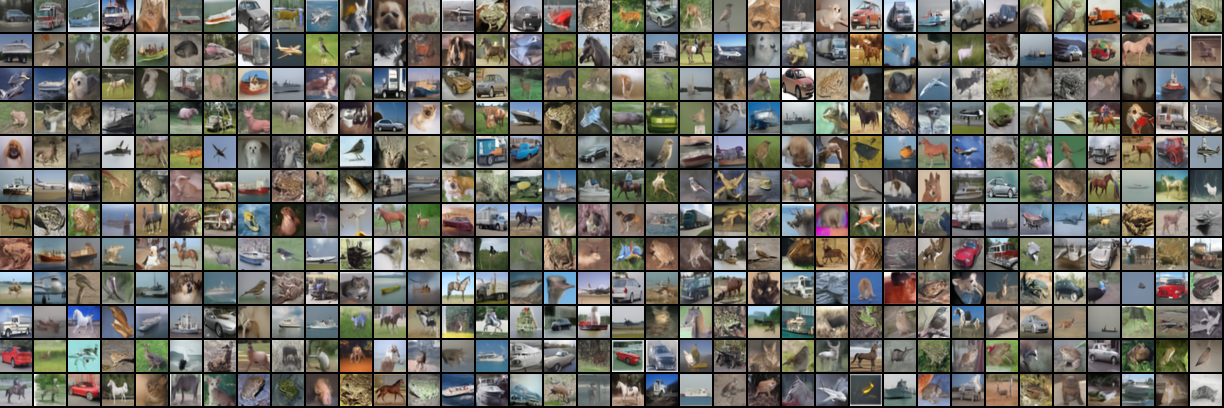}
        \caption{CUD Runge-Kutta 23 with 4 steps (FID=12.23)}
    \end{subfigure}
    \begin{subfigure}[!b]{\textwidth}
        \includegraphics[width=\textwidth]{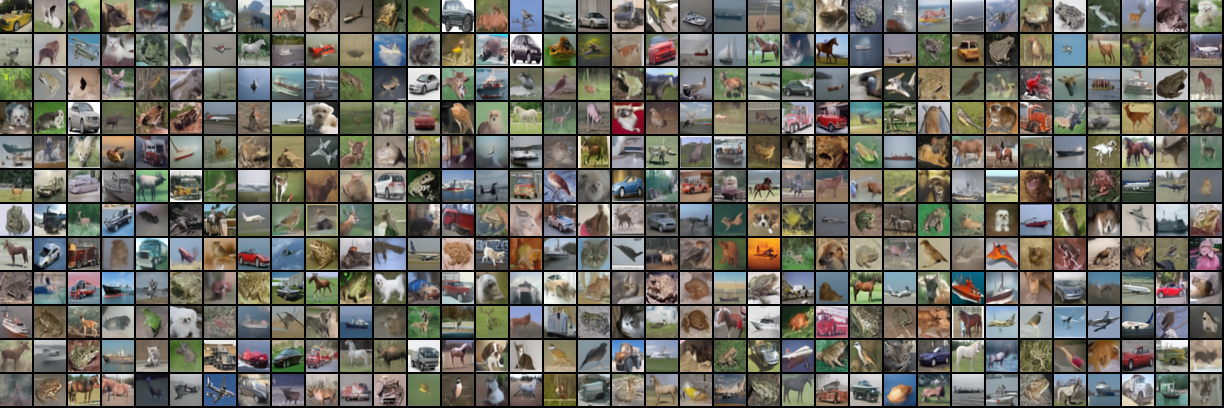}
        \caption{CUD Runge-Kutta 34 with 4 steps (FID=9.45)}
    \end{subfigure}
    \caption{Synthetic images sampled from CIFAR-10 $32\times 32$.}
    \label{fig:cifar10_full}
\end{figure}
\begin{figure}[!b]
    \centering
    \begin{subfigure}[!b]{\textwidth}
        \includegraphics[width=\textwidth]{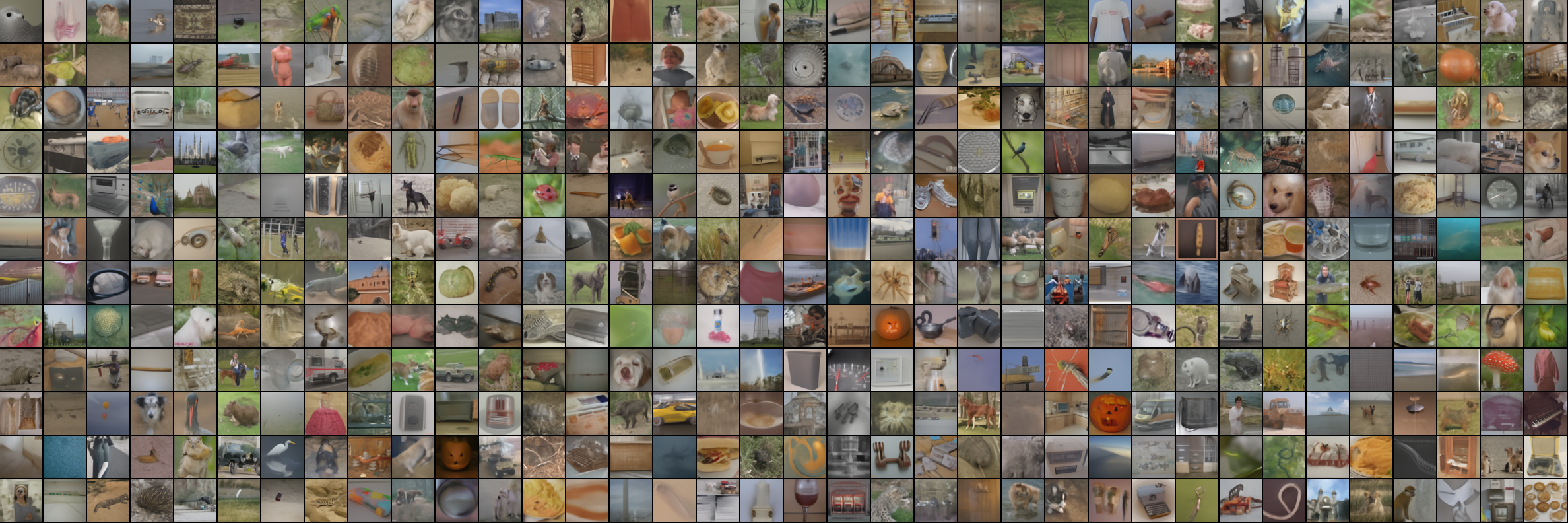}
        \caption{CUD Runge-Kutta 12 with 4 steps (Euler, FID=18.84)}
    \end{subfigure}
    \begin{subfigure}[!b]{\textwidth}
        \includegraphics[width=\textwidth]{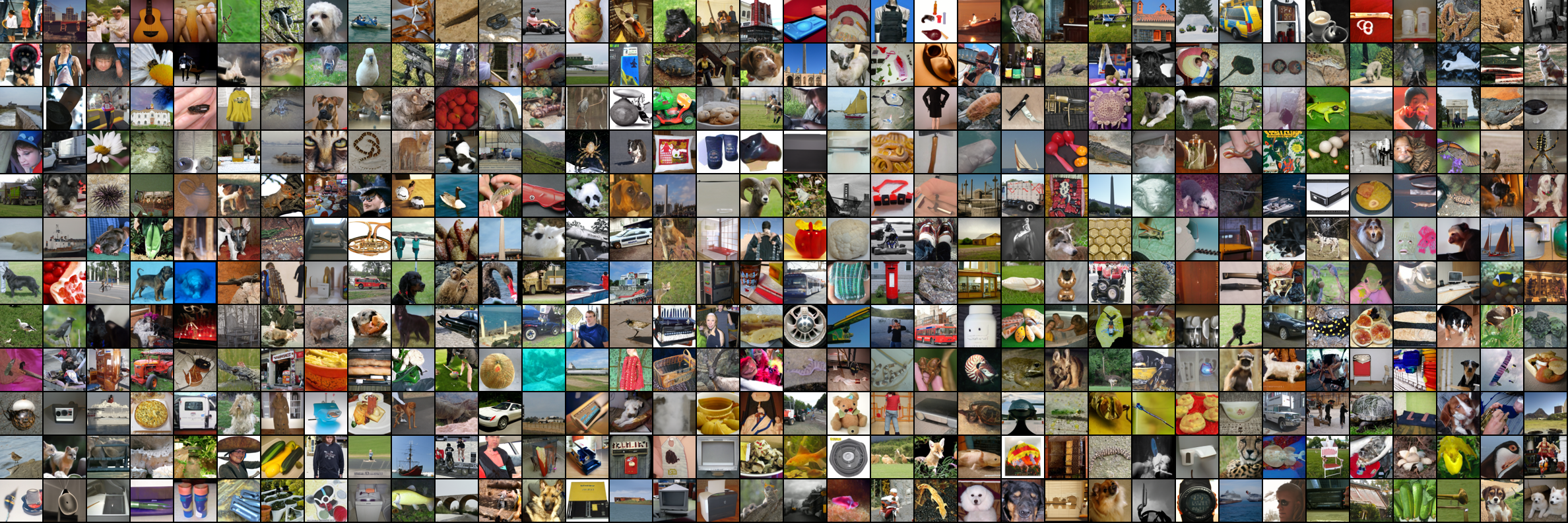}
        \caption{CUD Runge-Kutta 12 with 16 steps (Euler, FID=5.05)}
    \end{subfigure}
    \begin{subfigure}[!b]{\textwidth}
        \includegraphics[width=\textwidth]{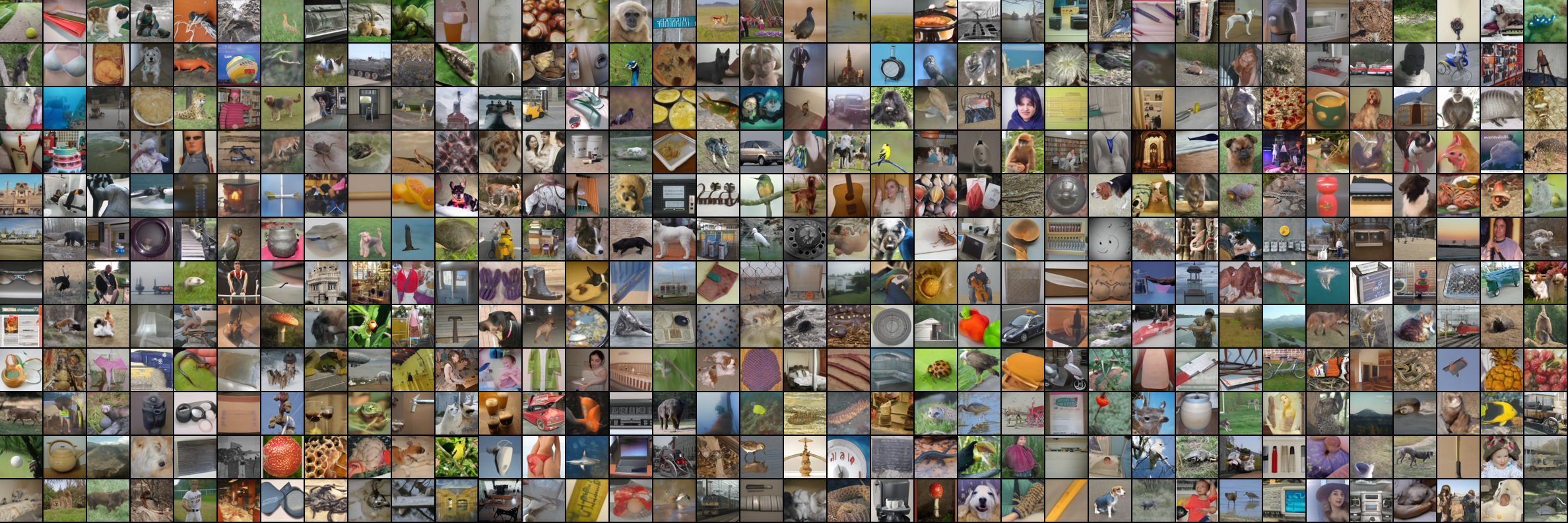}
        \caption{CUD Runge-Kutta 12 with 8 steps (Euler, FID=7.75)}
    \end{subfigure}
    \begin{subfigure}[!b]{\textwidth}
        \includegraphics[width=\textwidth]{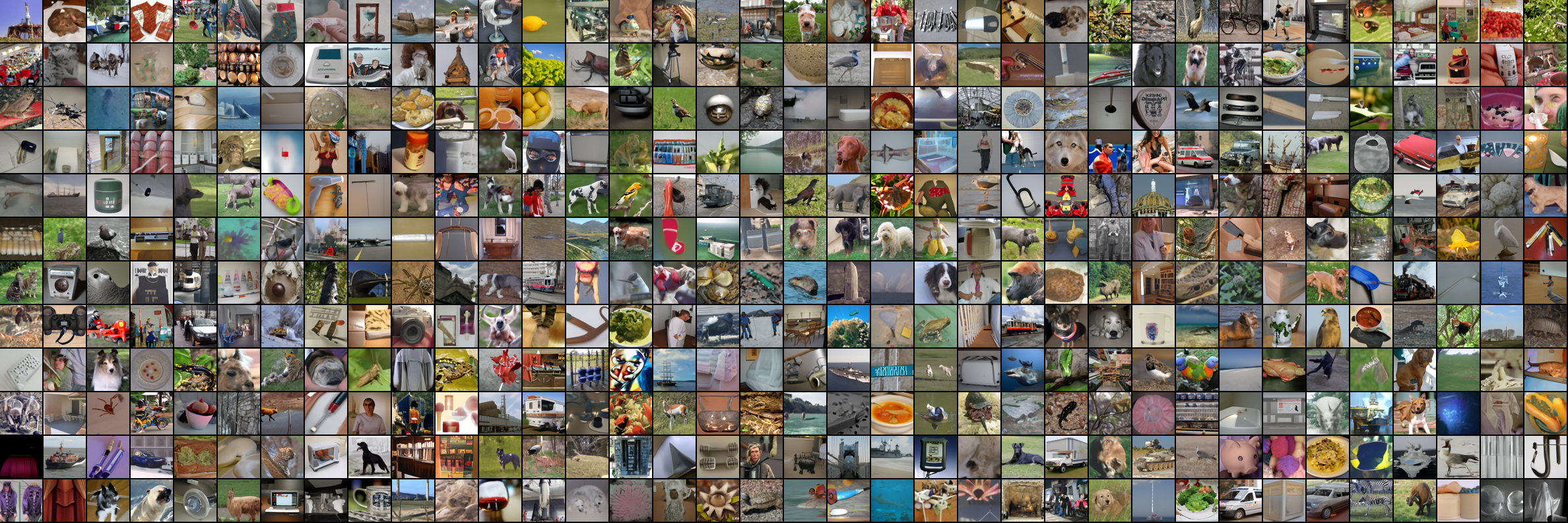}
        \caption{CUD Runge-Kutta 12 with 6 steps (DPM-Solver-2, FID=7.68)}
    \end{subfigure}
    \caption{Synthetic images sampled from ImageNet-1k $64\times 64$.}
    \label{fig:imagenet_64_full}
\end{figure}

\end{document}